\documentclass[a4paper,fleqn]{cas-sc}

\usepackage[authoryear]{natbib}
\usepackage{amsmath}
\usepackage{graphicx}
\usepackage{float}
\usepackage{placeins}
\usepackage{lipsum}
\usepackage{soul,color}
\usepackage{pdflscape}
\usepackage{rotating}
\usepackage{makecell}
\usepackage{placeins}
\usepackage{float}
\usepackage{booktabs}
\usepackage{caption}
\usepackage{graphicx}
\usepackage{amsthm} 

\newtheorem{assumption}{Assumption}
\newtheorem{remark}{Remark}
\usepackage{enumitem}

\usepackage{algorithm}
\usepackage{algpseudocode}
\usepackage{booktabs}   
\usepackage{multirow}   
\usepackage{siunitx}   

\allowdisplaybreaks

\usepackage{ulem}

\usepackage{hyperref}
\usepackage{subcaption}  

\usepackage{color}
\usepackage[english]{babel}
\usepackage{amsthm}

\usepackage{etoolbox}
\usepackage{xspace}

\numberwithin{equation}{section}

\usepackage[pagewise]{lineno}

\def\tsc#1{\csdef{#1}{\textsc{\lowercase{#1}}\xspace}}
\tsc{WGM}
\tsc{QE}
\tsc{EP}
\tsc{PMS}
\tsc{BEC}
\tsc{DE}

\begin{document}
\let\WriteBookmarks\relax
\def\floatpagepagefraction{1}
\def\textpagefraction{.001}
\shorttitle{PMA-Diffusion: A Physics-guided Mask-Aware Diffusion Framework for TSE from Sparse Observations}
\shortauthors{L. Liu, Z. Jin, and S. Choi}

\title [mode = title]{PMA-Diffusion: A Physics-guided Mask-Aware Diffusion Framework for Traffic State Estimation from Sparse Observations}

\author[1]{Lindong Liu}[orcid=0009-0001-7071-7002]
\ead{liu03035@umn.edu}
\credit{Conceptualization, Data curation, Formal analysis, Investigation, Methodology, Software, Validation, Visualization, Writing – original draft}

\author[2]{Zhixiong Jin}[orcid=0000-0002-1370-781X]
\ead{zhixiong.jin@univ-eiffel.fr}
\credit{Conceptualization of this study, Validation Writing - review and editing}

\author[1]{Seongjin Choi}[orcid=0000-0001-7140-537X]
\ead{chois@umn.edu}
\cormark[1]
\credit{Conceptualization, Data curation, Formal analysis, Funding acquisition, Investigation, Methodology, Project administration, Supervision, Writing – original draft, Writing – review \& editing}

\address[1]{Department of Civil, Environmental, and Geo- Engineering, University of Minnesota, 500 Pillsbury Dr. SE, Minneapolis, MN 55455, USA}
\address[2]{Univ. Gustave Eiffel, ENTPE, EMob-Lab, Lyon, F-69675, France}

\cortext[cor1]{Corresponding author}

\begin{abstract}
High-resolution highway traffic state information is essential for Intelligent Transportation Systems, but typical traffic data acquired from loop detectors and probe vehicles are often too sparse and noisy to capture the detailed dynamics of traffic flow. We propose PMA-Diffusion, a physics-guided mask-aware diffusion framework that reconstructs unobserved highway speed fields from sparse, incomplete observations. Our approach trains a diffusion prior directly on sparsely observed speed fields using two mask-aware training strategies: Single-Mask and Double-Mask. At the inference phase, the physics-guided posterior sampler alternates reverse-diffusion updates, observation projection, and physics-guided projection based on adaptive anisotropic smoothing to reconstruct the missing speed fields. The proposed framework is tested on the I-24 MOTION dataset with varying visibility ratios. Even under severe sparsity, with only 5\% visibility, PMA-Diffusion outperforms other baselines across three reconstruction error metrics. Furthermore, PMA-diffusion trained with sparse observation nearly matches the performance of the baseline model trained on fully observed speed fields. The results indicate that combining mask-aware diffusion priors with a physics-guided posterior sampler provides a reliable and flexible solution for traffic state estimation under realistic sensing sparsity.
\end{abstract}




\maketitle
\section{Introduction}\label{sec:intro}
High-resolution traffic state information is essential for both transportation operations and research. It supports Advanced Transportation Management Systems (e.g., ramp metering \citep{belletti2017expert}, variable speed limits \citep{zhang2025real}, coordinated traffic signal control \citep{ahmed2019real}, and incident detection \citep{coursey2025real}) and Advanced Traveler Information Systems (e.g., travel time reliability measures \citep{liu2004uncovering}, dynamic route planning \citep{ahmed2016prediction}, and traveler advisories \citep{liebig2017dynamic, mori2015review}). It also provides transportation researchers with detailed insights into various traffic phenomena, such as congestion dynamics, bottleneck formation, and the propagation of traffic waves \citep{ji2026scalable}. However, obtaining such high-resolution observations at scale remains challenging. In practice, most highway monitoring systems highly rely on sparse loop detectors (installed approximately half-mile intervals \citep{liu2006detector, MnDOT2019ITS}), which provide consistent but location-limited measurements, or probe vehicles that provide broader spatial coverage but suffer from low and uneven penetration rates (typically 2-5\% penetration rate \citep{seo2017traffic,herrera2010evaluation}). Emerging sensing platforms, such as Unmanned Aerial Vehicles (UAV)-based traffic monitoring (e.g., pNEUMA \citep{barmpounakis2020new}, Songdo traffic data \citep{fonod2025advanced}) and large-scale multi-camera systems (I-24 MOTION testbed \citep{gloudemans202324}), can provide near-complete and fine-grained observations but are constrained by battery capacity, altitude restrictions, and privacy regulations, and deployment cost \citep{butilua2022urban}. As a result, it is difficult to fully observe traffic states at high resolution, making it necessary to develop methods that can reconstruct them from sparse and incomplete measurements.

Traffic State Estimation (TSE) addresses this problem by reconstructing high-resolution traffic data from sparse and heterogeneous measurements \citep{wang2023low}. Existing methods are often grouped into model-driven and data-driven approaches \citep{seo2017traffic}. Model-driven methods rely on macroscopic traffic-flow theory, such as the Lighthill-Whitham-Richards (LWR) model and its extensions \citep{lighthill1955kinematic,coifman2002estimating}. These models are interpretable and can extrapolate traffic states from relatively few measurements, but their assumptions, such as homogeneous traffic conditions and stable wave speeds, often do not hold in the real world \citep{shi2021physics}. Data-driven methods instead learn traffic states directly from historical data and can flexibly capture nonlinear spatial-temporal relationships without specifying a detailed traffic-flow model. Even though recent deep learning models can capture complex spatial-temporal dependencies, they typically require high-visibility data and corresponding full ground truth. When the available data are sparse, or when the model is conditioned only on limited observations, the reconstructions can violate basic traffic principles \citep{thodi2022incorporating}.

Recent work on Physics-Informed Deep Learning (PIDL) incorporates traffic theory into the learning objective \citep{shi2021physics, di2023physics}. Most PIDL methods use the conservation laws or the fundamental diagram relations by penalizing their Partial Differential Equation (PDE) residuals or adding physics-based regularization terms to the loss \citep{raissi2019physics}. When traffic states are fully or densely observed, these physics terms help stabilize training and lead to more reliable estimations \citep{yuan2021macroscopic, wu2024traffic}. However, such full-observation conditions are rare in practice due to sparse and limited sensor coverage. In addition, physics terms are usually encoded directly in the loss under a fixed detector layout and masking pattern. In other words, when the sensor configuration changes at deployment, the training-time masks no longer match the new observation measurement \citep{xue2024network}.

In summary, current TSE methods share three fundamental limitations. First, most learning-based models \textbf{require fully observed traffic states} for model training. For example, previous Variational Autoencoder (VAE) and Diffusion-based generative models rely on complete spatiotemporal speed fields for training \citep{boquet2020variational, lei2024conditional, lu2025diffusion, zhang2025tsgdiff}, and Generative Adversarial Network (GAN) based models are trained by matching sparse observations to their corresponding full ground-truth speed data \citep{zhang2022tsr, mo2022trafficflowgan}. Gaussian process (GP)-based approaches provide a principled probabilistic alternative \citep{yuan2021macroscopic, wu2024traffic, storm2022efficient}, but they depend on predefined physics-informed kernels or relatively dense observations for calibration.
Second, most existing methods \textbf{produce only a single point estimate}, instead of modeling the uncertainty and full posterior distribution of the complete speed field. In sparse sensing regimes, multiple traffic states can be consistent with the same observations, so representing the posterior distribution is important for uncertainty quantification and downstream decision making. This limitation appears in classical reconstruction methods \citep{treiber2002reconstructing, treiber2011reconstructing, tak2016data, wang2023low}, in the majority of deep regression models \citep{rempe2022estimation, liu2021learning}, and in many PIDL approaches \citep{shi2021physics, huang2022physics, zhang2024physics, shi2023physics, abewickrema2026physics, thodi2022incorporating}. GP-based methods \textit{do} provide predictive variances, but these variances are induced by the chosen kernel rather than arising from a learned generative model of traffic dynamics.
Third, current PIDL frameworks typically encode the detector layout and observation mask directly into the loss function. Since these physics penalties are tied to the specific observation operator used during training \citep{huang2020physics, shi2023physics}, the learned model \textbf{becomes sensitive to changes in detector configuration}.

In this paper, we propose \textbf{PMA-Diffusion}, a Physics-guided Mask-Aware Diffusion framework for reconstructing high-resolution traffic speed fields from sparse loop detectors and probe vehicle data. \textbf{PMA-Diffusion} addresses the aforementioned three key limitations of existing TSE methods by considering TSE as an inverse problem and decoupling generative prior training and posterior sampling. In the inverse problem formulation, the goal is to recover the unobserved traffic state ($\mathbf{x}$) from sparse and noisy observation ($\mathbf y = \mathcal A(\mathbf x) + \boldsymbol\eta$), where $\mathcal{A}$ denotes the measurement operator (e.g., observation mask) and $\boldsymbol\eta$ denotes measurement noise.
PMA-Diffusion first learns a prior distribution ($p(\mathbf{x})$) of high-resolution traffic state directly from sparse observations ($\mathbf{y}$) using Single-mask and Double-mask training strategies. The Single-mask strategy follows the actual sensor visibility pattern, whereas the Double-mask strategy adds an auxiliary mask so that even always-visible locations are occasionally treated as unobserved during training. This removes the dependence on full observation for training. Then, to produce the distribution over all plausible high-resolution traffic states given observation, we design a physics-guided posterior sampling process ($p(\mathbf{x\mid y}) \propto p(\mathbf{x})\cdot p(\mathbf{y\mid x})$) \citep{lugmayr2022repaint,chung2024deep}. Specifically, during posterior sampling, PMA-Diffusion alternates reverse-diffusion updates with two projection steps: an observation projection that injects the measured values ($p(\mathbf{\mathbf{y\mid x}})$) back into the trajectory, and a physics-guided projection that enforces basic kinematic-wave structure on the unobserved regions. Since these projections are applied during inference rather than encoded into the loss, the framework can be used with arbitrary observation masks over highway speed fields. 

This paper is organized as follows. Section~\ref{sec:back} reviews related work on traffic state estimation. Section~\ref{sec:meth} describes the methodology of this paper. A detailed problem formulation is presented in Section~\ref{sec:probs}, the proposed framework of PMA-Diffusion is presented in Section~\ref{sec:ambient} and Section~\ref{sec:sampler}. Section~\ref{sec:experiments} describes the experiments and evaluations. Section~\ref{sec:conclu} provides conclusions.

\section{Background}\label{sec:back}
In this section, we review three lines of work related to our proposed framework: (i) Model-Driven and Data-Driven Approaches for TSE, (ii) Physics-Informed Deep Learning (PIDL) for TSE, and (iii) Diffusion-based Approaches for Inverse Problems and TSE.

\subsection{Model-Driven and Data-Driven Approaches for TSE}
Early work on TSE was based on traffic flow theory and aimed to reconstruct complete traffic states from sparse sensor measurements. The first group of methods directly relies on macroscopic traffic flow equations, most commonly the LWR model \citep{lighthill1955kinematic, richards1956shock} and its discretized form, the Cell Transmission Model (CTM) \citep{daganzo1995cell}. These methods solve the traffic flow PDE and use boundary conditions to estimate the traffic states. For example, \cite{coifman2002estimating} estimated traffic states using data from a single double-loop detector based on the fundamental diagram relationship. \cite{seo2015probe} incorporated probe vehicle spacing data into the LWR conservation law to infer density and flow in Eulerian coordinates, enabling state estimation under low probe penetration rates. 
In addition to PDE-based formulations, \cite{treiber2003adaptive} proposed the Adaptive Smoothing Method (ASM), which reconstructs spatiotemporal speed fields by applying different smoothing strengths along the characteristic directions of free-flow and congested traffic waves \citep{treiber2002reconstructing,treiber2011reconstructing}. 
A second group of methods rewrites the traffic flow PDE as a dynamic model with a corresponding observation model, allowing recursive estimation through filtering techniques. Following this idea, \cite{wang2005real} discretized a second-order traffic flow model \citep{papageorgiou1990modelling} into a nonlinear state-space form and applied the Extended Kalman Filter (EKF) to integrate loop detector data for traffic state estimation on highway. Subsequent studies improved performance under nonlinearities and high-dimensional state representations by using the Unscented Kalman Filter (UKF) and the Ensemble Kalman Filter (EnKF) \citep{mihaylova2006unscented,work2008ensemble}. Despite their data efficiency and physical interpretability, model-driven methods are highly sensitive to model structure and parameter choices. Offline calibration of fundamental-diagram shape and wave speeds requires large volumes of high-quality historical data and is subject to drift as demand patterns, driver behavior, and control strategies evolve \citep{yuan2021macroscopic,di2023physics}. 

In contrast, data-driven methods infer traffic states directly from historical data \citep{seo2017traffic}. Early data-driven studies mainly adopted traditional statistical learning techniques to impute missing traffic measurements. For example, \cite{zhong2004estimation} employed Auto Regressive Integrated Moving Average (ARIMA) to reconstruct missing data. \cite{ni2005markov} further addressed uncertainty in missing observations by using Bayesian Networks.
Researchers then introduced non-parametric methods that avoid strong functional assumptions. Kernel regression smooths inconsistent measurements using locally weighted averages \citep{yin2012imputing}, and k-nearest neighbors reconstruct speed or flow by identifying similar historical patterns \citep{tak2016data}. Another line of work uses the low-rank structure of traffic states in space and time: \cite{coric2012traffic} reconstructed high-resolution speed fields from aggregated data using signal-reconstruction techniques, while \cite{wang2023low} used Hankel tensor decomposition to capture temporal continuity and impute missing values. These approaches are effective in capturing recurrent patterns, but their accuracy decreases under abrupt traffic condition changes or unreliable sensor measurements. More recently, deep learning has expanded the scope of data-driven TSE. Generative adversarial networks (GANs) have been applied to quantify uncertainty and reconstruct missing traffic states from partial observations \citep{mo2022quantifying}. Normalizing flows provide flexible probabilistic models for complex traffic state distributions \citep{huang2023bridging}, while variational autoencoders (VAEs) embed traffic states into low-dimensional latent spaces for imputation and anomaly detection \citep{boquet2020variational}. Despite these advances, deep learning models remain difficult to interpret \citep{shi2021physics} and generally require large amounts of representative data to achieve stable performance over regions with long-term missing observations. Beyond these parametric and deep learning approaches, Gaussian process (GP) models provide a nonparametric probabilistic approach to TSE. Several recent works introduce GP priors whose covariance structure reflects macroscopic traffic-flow properties. \citet{yuan2021macroscopic} proposed a physics-regularized Gaussian process in which the kernel is constructed such that sampled trajectories remain close to solutions of the kinematic-wave model. \citet{wu2024traffic} developed anisotropic Gaussian processes with different propagation speeds in free-flow and congested states, capturing the directional nature of traffic waves. GP-based methods offer uncertainty estimates and can incorporate simple forms of traffic physics through kernel design. However, they require relatively dense observations to calibrate kernel hyperparameters, and the resulting uncertainty reflects interpolation error rather than the posterior variability arising from structural sparsity.

Recent work has applied diffusion models to traffic state estimation under different observation settings. Under fully observed or densely observed data, diffusion models have been used to learn spatiotemporal traffic patterns directly. For example, \citet{lei2024conditional} trained a conditional diffusion model on complete speed fields and then conditioned on partial values to estimate traffic states at uninstrumented locations. Similarly, \citet{lu2025diffusion} introduced a multi-scale diffusion structure that exploits correlations between coarse and fine resolutions to impute missing detector data, assuming that full-resolution targets are available during training.
A second set of studies considers heterogeneous sensing environments. \citet{zhang2025tsgdiff} proposed a graph-based diffusion framework that integrates loop detectors, probe vehicles, and contextual information, and \citet{lyu2025diffusion} formulated a diffusion-based expectation-maximization strategy to jointly handle noise and missing entries by treating the clean traffic field as a latent variable. These works demonstrate that diffusion priors can capture variability in traffic conditions and can operate with multiple data sources when the training set is sufficiently complete. High-resolution trajectory studies also highlight the importance of dense fields for downstream analysis. For instance, \citet{ji2026scalable} used nearly complete vehicle trajectories from the I-24 MOTION testbed to quantify stop-and-go wave characteristics. While this line of work does not address reconstruction from sparse loop-probe networks, it underscores the level of detail required for dynamical analysis and further motivates methods capable of recovering such fields from limited observations.

Overall, existing TSE methods show a clear trade-off between using physical models and relying on data. Model-driven approaches are interpretable and work well when the traffic model and its parameters are accurate, but their performance degrades when the assumed model does not match actual traffic conditions. In contrast, data-driven approaches can capture rich spatiotemporal relationships without specifying a traffic model, but they usually rely on densely observed or well-imputed training data and provide limited physical guarantees when observations are sparse or heterogeneous.

\subsection{Physics-Informed Deep Learning (PIDL) for TSE}
Physics-Informed Deep Learning (PIDL) incorporates traffic-flow theory into neural networks by adding constraints derived from conservation laws or fundamental diagram relations to the training objective \citep{raissi2019physics,di2023physics}. In highway traffic state estimation, \cite{huang2020physics} applied this idea to first-order LWR and CTM formulations, developed a Physics-Informed Neural Network (PINN) framework based on the LWR conservation law and Greenshields’ fundamental diagram. \cite{shi2021physics} extended the PINN framework to a second order Aw-Rascle-Zhang (ARZ) model by penalizing continuity and momentum residuals in the loss. Based on these PINN based method, \cite{zhang2024physics} represented the discretized traffic flow model as a computational graph and treated fundamental diagram parameters as trainable variables. \cite{wang2024knowledge} developed a stochastic physics-informed model that treats demand and model parameters as random variables, allowing the method to estimate traffic states and parameters together with their uncertainty. For signalized arterial roads, \cite{abewickrema2026physics} developed a phase-aware PIDL framework that incorporates conservation constraints and signal-phase information into the loss to estimate traffic state.
In parallel, several works have incorporated traffic-flow physics into nonparametric probabilistic models that are not deep networks but play a similar regularizing role \citep{di2023physics}. \cite{yuan2021macroscopic} proposed a physics-regularized Gaussian process model in which the covariance structure is designed so that sampled trajectories remain close to solutions of a macroscopic kinematic-wave model . \cite{wu2024traffic} developed anisotropic Gaussian processes whose kernels encode different propagation speeds in free-flow and congested states, effectively embedding kinematic-wave anisotropy into the prior. At the network scale, \cite{xue2024network} designed a graph neural network for traffic state imputation and added penalties that discourage violations of a network macroscopic fundamental diagram, so that the imputed states are consistent with aggregate network flow theory.

These PIDL and physics-regularized methods show that embedding macroscopic flow constraints can improve robustness and interpretability when complete or densely observed traffic states are available, or when density and flow can be reliably reconstructed. However, they also have two important limitations relative to our setting. First, detector layout and observation mask are typically encoded directly in the training loss, tightly coupled to a particular sensor layout and masking pattern. When detector locations or probe penetration rates change at deployment, the training-time loss no longer matches the actual measurement operator, and performance may degrade \citep{di2023physics,xue2024network}. Second, most existing PIDL approaches require access to complete traffic state fields during training. In practice, such complete fields are rarely available, and loop detector and probe vehicle networks provide only sparse and irregular samples. As a result, standard PIDL methods are not appropriate to apply directly in incompletely observed regimes.

\subsection{Traffic State Estimation as an Inverse Problem}
\subsubsection{Inverse problems and the position of TSE}
Inverse problems aim to recover an unknown signal or field $\mathbf x$ from indirect, noisy measurements $\mathbf y$, typically modeled as
$\mathbf y = \mathcal A(\mathbf x) + \boldsymbol\eta$, where $\mathcal A$ is a known or parametrized forward operator and $\boldsymbol\eta$ denotes measurement noise \citep{tarantola2005inverse, ribes2008linear}. Many problems in imaging, geophysics, and remote sensing fit this form and are ill-posed in the sense of non-uniqueness or instability, so meaningful solutions require regularization or Bayesian priors on $\mathbf x$ \citep{barrett2013foundations, daubechies2004iterative, donoho2006compressed}. A standard distinction is between linear and nonlinear problems, depending on whether $\mathcal A$ is linear, and between non-blind and blind formulations, depending on whether the operator and noise statistics are known \citep{ribes2008linear}. From a Bayesian viewpoint, both cases can be written as inference on a posterior density
\begin{equation}
    p(\mathbf x \mid  \mathbf y) \;\propto\; p(\mathbf y \mid  \mathbf x)\,p(\mathbf x),
\end{equation}
where $p(\mathbf x)$ encodes prior information about the unknown field and $p(\mathbf y \mid  \mathbf x)$ is the likelihood induced by $\mathcal A$ and $\boldsymbol\eta$.

Classical approaches impose analytic regularizers such as $\ell_2$, $\ell_1$, or total variation penalties and solve the resulting optimization problem \citep{daubechies2004iterative, donoho2006compressed}. More recent methods use plug-and-play denoisers or regularization-by-denoising strategies, which combine a physics-based forward model with a learned image prior inside an iterative algorithm \citep{venkatakrishnan2013plug, romano2017little, kamilov2023plug}. Deep generative models extend this idea by learning expressive priors $p(\mathbf x)$ from data and using them within posterior sampling or variational strategies for linear and mildly nonlinear inverse problems \citep{song2019generative, ho2020denoising, kawar2022denoising, chung2023diffusion}.

Traffic state estimation fits naturally into this framework. In the TSE setting, we let $\mathbf x$ denote the full speed field on a discretized highway corridor and let $\mathbf y$ collect sparse, noisy measurements from loop detectors, probe vehicles, and camera systems. The observation process can be represented as a linear operator that applies masking, spatial sampling, and local averaging to $\mathbf x$, followed by additive noise. When sensor locations, aggregation intervals, and basic calibration are known, traffic state estimation corresponds to a large-scale, non-blind linear inverse problem with highly structured missing data. If detection rates, outages, or miscalibration are uncertain, the problem becomes semi-blind and more ill-conditioned. As we mentioned in previous subsection, existing traffic state estimation methods usually apply Kalman-type filters, PDE-based reconstruction methods, or deep learning models conditioned on partial observations. These approaches estimate states directly and do not learn a reusable generative prior over full speed fields that can be combined with different observation mask.

In this work, we formulate highway traffic state estimation as a Bayesian inverse problem with an unconditional prior $p_\theta(\mathbf V)$ over complete speed fields and a likelihood that encodes the mask-structured measurement operator. The prior is learned from historical data and does not depend on a particular detector layout or probe penetration rate. During inference, we combine $p_\theta(\mathbf V)$ with a masked Gaussian likelihood and a physics-motivated projector inside a diffusion-based sampler. This formulation follows recent developments in diffusion-based inverse problems in imaging, but is adapted to the linear, mask-structured operators and macroscopic flow constraints that arise in highway monitoring \citep{song2021solving, chung2022score, kawar2022denoising, chung2023diffusion}. It provides a common framework in which a single prior can be reused under different sensor configurations while physics enters through modular projection rather than through the training loss.

\subsubsection{Diffusion models as generative priors for inverse problems and TSE}
Diffusion and score-based models learn data distributions by gradually corrupting a clean sample $\mathbf x_0$ with Gaussian noise and training a neural network to approximate the reverse denoising process \citep{ho2020denoising, song2020score}. In the variance-preserving setting, the forward corruption at time $t$ has the closed-form expression; a detailed derivation and discussion can be found in \citet{song2020score}.
\begin{equation}
    \mathbf x_t = \sqrt{\alpha(t)}\,\mathbf x_0
    + \sqrt{1-\alpha(t)}\,\boldsymbol\varepsilon,
    \qquad 
    \boldsymbol\varepsilon\sim\mathcal N(\mathbf 0,\mathbf I),
\end{equation}
where $\alpha(t)$ is a decreasing function. A neural network $\varepsilon_\theta$ is trained to predict the injected noise through the denoising objective
\begin{equation}
    \mathcal L(\theta)
    = \mathbb E\!\left[\|\boldsymbol\varepsilon-\varepsilon_\theta(\mathbf x_t,t)\|^2\right],
\end{equation}
which provides a score approximation
\begin{equation}
    \nabla_{\mathbf x_t}\log p_t(\mathbf x_t)
    \approx -\frac{1}{\sigma_t}\,\varepsilon_\theta(\mathbf x_t,t).
\end{equation}
After discretizing the reverse dynamics, the model defines a Markov chain whose stationary distribution approximates the data distribution. The learned model therefore acts as an unconditional prior distribution, $p_\theta(\mathbf x)$, and sometimes referred to as a \textit{generative prior} or a \textit{diffusion prior}.

Generative priors can be combined with known forward operators to solve inverse problems. A typical solver alternates (i) a reverse-diffusion update, which moves the current iterate toward high-density regions under $p_\theta(\mathbf x)$, and (ii) a data-consistency step that enforces agreement with the observations. RePaint performs inpainting by interleaving reverse steps with occasional forward re-noising and exact replacement on the observed pixels \citep{lugmayr2022repaint}. DDRM derives closed-form conditional updates for linear Gaussian operators \citep{kawar2022denoising}. Diffusion Posterior Sampling (DPS) incorporates likelihood gradients or approximate pseudoinverses to guide sampling for a range of linear problems, including MRI reconstruction \citep{song2021solving, chung2022score, chung2023diffusion}. Recent theoretical work studies these algorithms as approximations to Bayesian posterior sampling in high-dimensional inverse problems.

\section{Methodology}\label{sec:meth}
In this section, we present the details of PMA-Diffusion. The goal is to reconstruct high-resolution traffic speed fields from sparse and heterogeneous measurements. The method has two main components: a mask-aware diffusion prior and a physics-guided sampling strategy.
The first stage aims to train the generative prior $p_\theta(V)$ directly from incomplete observations. After the training stage, we then use the trained diffusion prior in the Physics-Guided posterior sampling stage to inject measurements during the reverse process of the diffusion model through the resampling strategy and enforce traffic constraints through a physics-guided projection operator.

\begin{figure}[width=.99\linewidth,cols=4,pos=t]
    \centering
    \includegraphics[
        width=\linewidth,
        trim=15mm 65mm 15mm 58mm,
        clip
    ]{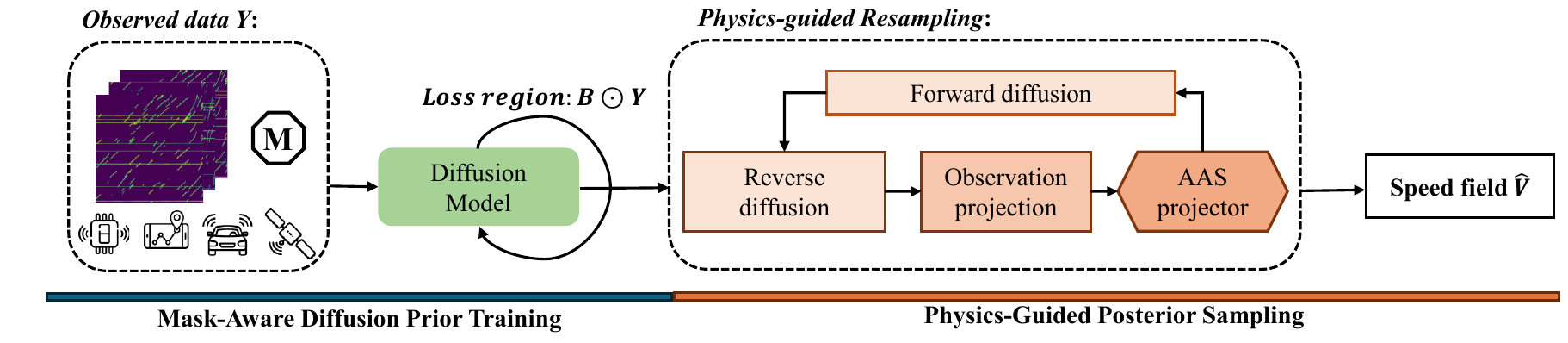}
    \caption{Overall pipeline of the proposed PMA-Diffusion framework.}
    \label{fig:framework}
\end{figure}

Figure~\ref{fig:framework} summarizes the overall pipeline of the proposed PMA-Diffusion Framework. The following subsections formalize the problem setting, present the mask-aware diffusion training strategy, and describe the physics-guided sampling procedure.

\subsection{Problem Setting}
\label{sec:probs}
To formalize the reconstruction task, we discretize the highway corridor into $S$ spatial steps and $T$ time steps. Let $\mathbf V \in [0,1]^{S\times T}$ denote the latent (noise-free) normalized speed field, obtained by dividing physical speeds by a fixed $v_{\max}$ so that each entry satisfies $V_{s,t}\in[0,1]$. Only a small fraction of entries are observed; the remainder must be inferred.
\begin{figure}[width=.99\linewidth,cols=4,pos=t]
    \centering
    \includegraphics[width=0.8\linewidth,
        trim=15mm 32mm 15mm 20mm,
        clip]{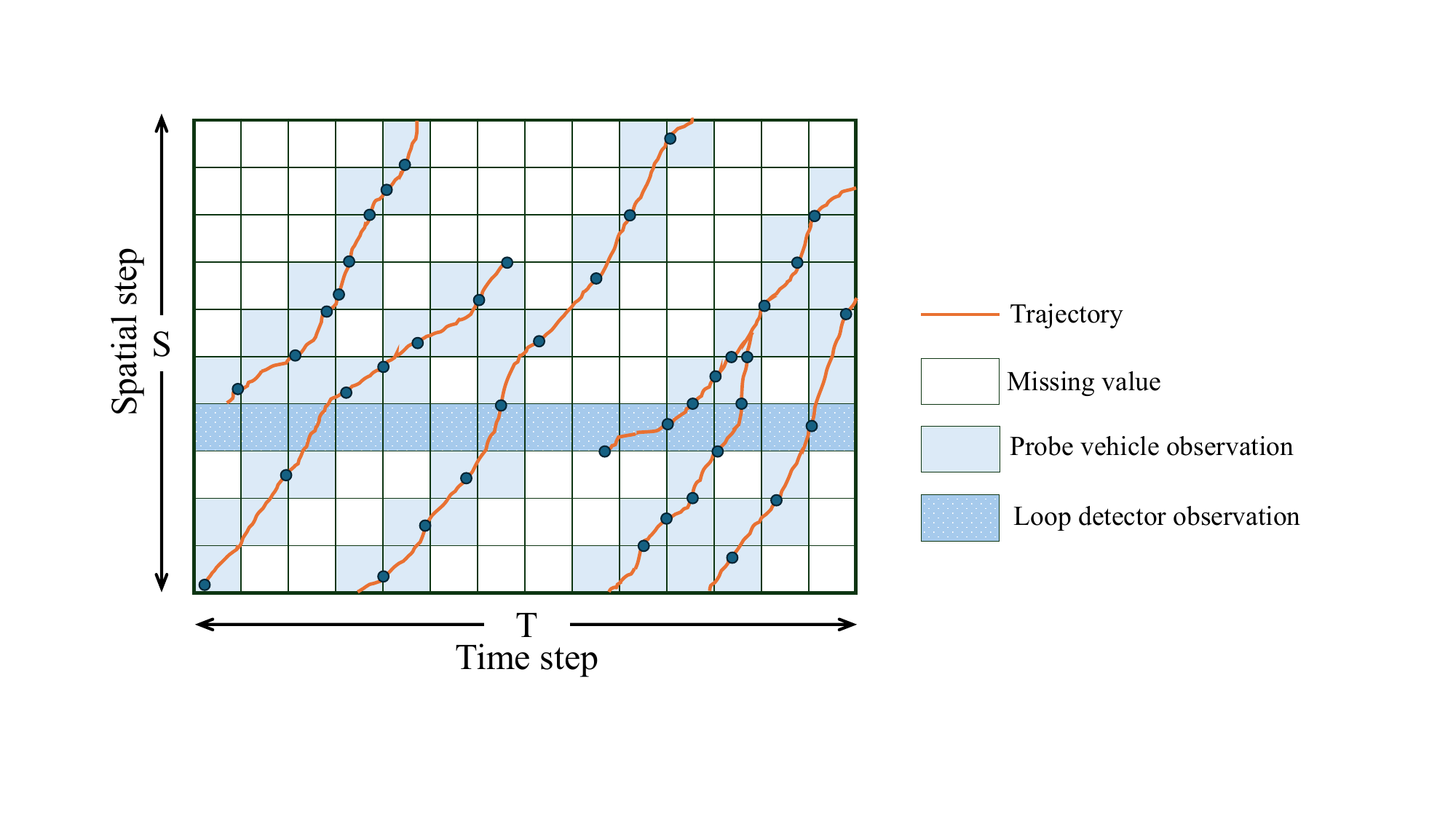}
    \caption{Matrix representation of traffic state variables collected from probe vehicles and loop detectors.}
    \label{fig:speed field}
\end{figure}

\subsubsection{Observation measurement} 
Our observed data come from two asynchronous sources: loop detectors and probe vehicles. Their spatio-temporal coverage is illustrated in Figure~\ref{fig:speed field}.
\begin{itemize}
    \item \textbf{Loop detectors.} Loop detectors are fixed-point sensors installed at selected highway segments. Their spatial configuration is time-invariant and is represented by a binary mask $\mathbf M^{(\ell)}\in\{0,1\}^{S\times T}$, where $M^{(\ell)}_{s,t}=1$ indicates an observed entry. The related observation noise is modeled as $\boldsymbol\varepsilon^{(\ell)}\sim\mathcal N(0,\sigma_\ell^{2}\mathbf I)$.
    \item \textbf{Probe vehicles.} Probe vehicle data are generated from moving vehicles equipped with GPS or other communication devices. Unlike loop detectors, probe vehicle data exhibit a stochastic sampling pattern across space and time, represented by $\mathbf M^{(p)}\in\{0,1\}^{S\times T}$. The related observation noise is modeled as $\boldsymbol\varepsilon^{(p)}\sim\mathcal N(0,\sigma_p^{2}\mathbf I)$.
\end{itemize}
The combined observation mask is defined as $\mathbf M=\mathbf M^{(\ell)} \lor \mathbf M^{(p)}$. In typical highway monitoring scenarios, the visibility ratio is $\rho=\|\mathbf M\|_0/(ST)$, where $\|\mathbf M\|_0=\sum_{s,t} M_{s,t}$ is the number of observed entries, and we typically have $\rho\ll 1$.

Incorporating a possible GPS latency $\tau$, we write the observation operator $\mathcal O_\rho$ as
\begin{equation}
    \mathcal O_\rho(\mathbf V)=
    \mathbf M^{(\ell)} \odot \bigl(\mathbf V + b^{(\ell)} + \boldsymbol\varepsilon^{(\ell)}\bigr)
    +
    \mathbf M^{(p)} \odot \bigl(\mathsf S_{-\tau}\mathbf V + b^{(p)} + \boldsymbol\varepsilon^{(p)}\bigr),
    \label{eq:obs-general}
\end{equation}
where $\odot$ denotes the element-wise product, $b^{(\cdot)}$ denotes a sensor-specific bias field, and $\mathsf S_{-\tau}$ is a shift operator that moves the field backward in time by the GPS latency $\tau$, consistent with \citet{kim2014comparing}.

\subsubsection{Empirical evidence and assumptions}
Recent physics-informed and data-driven TSE studies typically assume that GPS latency ($\tau$) is negligible relative to the aggregation interval, and that low-frequency sensor biases $b^{(\cdot)}$ can be absorbed into the observation noise $\boldsymbol\varepsilon^{(\cdot)}$ without explicit modeling \citep{shi2021physics,wu2024traffic,wang2023low}. We follow the same practice and make the following working assumptions.

\begin{assumption}[Negligible latency]\label{asm:latency} 
We assume $\tau=0$, so probe vehicle data are effectively time-aligned with loop detector data.
\end{assumption} 

\begin{assumption}[Bounded low-frequency drift]\label{asm:drift}
The sensor bias field $b^{(\cdot)}$ is $L_b$-Lipschitz and bounded in amplitude, $\|b^{(\cdot)}\|_\infty \le \delta$. We absorb $b^{(\cdot)}$ into the noise term, which introduces at most $O(\delta)$ bias in the score used for training.
\end{assumption}

\begin{remark}
Consider the (single-source) conditional log-likelihood,
\begin{equation}
    \log p(Y\mid  V)=\log p_\eta\bigl(Y-V-b^{(\cdot)}\bigr).
\end{equation}
If $b^{(\cdot)}$ is Lipschitz-continuous and small in amplitude, a first-order Taylor expansion around $b^{(\cdot)}=0$ gives
\begin{equation}
    \log p_\eta(Y-V-b^{(\cdot)})=\log p_\eta(Y-V) - \nabla_y \log p_\eta(Y-V)^{\rm T} b^{(\cdot)} + O(\delta^2).
\end{equation}
Differentiating with respect to $V$ shows that the score is perturbed by at most $O(\delta)$ when we replace $(Y-V-b^{(\cdot)})$ with $(Y-V)$. This motivates treating the masked score estimator as approximately unbiased when $\delta$ is small.
\end{remark}

Under Assumptions~\ref{asm:latency}-\ref{asm:drift}, the observation model in Eq.~\eqref{eq:obs-general} simplifies to
\begin{equation}
    \mathbf Y \;=\;
    \mathbf M^{(\ell)}\odot\bigl(\mathbf V+\boldsymbol\varepsilon^{(\ell)}\bigr)
    \;+\;
    \mathbf M^{(p)}  \odot\bigl(\mathbf V+\boldsymbol\varepsilon^{(p)}\bigr).
    \label{eq:obs-simple}
\end{equation}

To make the subsequent masking analysis clear, we impose the following additional assumptions.
\begin{assumption}[Noise conditional independence]\label{asm:indep}
Given the latent speed field $\mathbf V$, the loop detector and probe vehicle noise fields are independent:
$\boldsymbol\varepsilon^{(\ell)} \perp\!\!\!\perp \boldsymbol\varepsilon^{(p)} \mid  \mathbf V$.
\end{assumption}
\begin{assumption}[Missing-at-random masking]\label{asm:mar}
Conditioned on $\mathbf V$, the masking process is independent of the observation noise:
\[
\mathbf M^{(\cdot)} \;\perp\!\!\!\perp\; \boldsymbol\varepsilon^{(\cdot)} \mid \mathbf V.
\]
\end{assumption}

Under these working assumptions, we can write the observation model in a compact form as
\begin{equation}
    \mathbf Y \;=\; \mathbf M \odot \bigl(\mathbf V + \boldsymbol\varepsilon\bigr), 
    \qquad \boldsymbol\varepsilon \sim\mathcal N(0,\sigma^{2}\mathbf I).
    \label{eq:obs-compact}
\end{equation}

\subsubsection{Bayesian posterior perspective}\label{sec:bayes_posterior}
Given $\mathbf{Y}$ and $\mathbf{M}$, our goal is to draw samples from the posterior distribution
\begin{equation}
    p(\mathbf V \mid  \mathbf Y)\;\propto\;
    \underbrace{\mathbf 1\!\bigl[\mathbf V\in\mathcal P_{\text{phys}}\bigr]}_{\text{physics surrogate}}
    \;\times\;
    \underbrace{p_\theta(\mathbf V)}_{\text{mask-aware diffusion prior}}
    \;\times\;
    \underbrace{\exp \!\Bigl[-\tfrac{1}{2\sigma^{2}}\bigl\|(\mathbf V\odot\mathbf M)-\mathbf Y\bigr\|_F^{2}\Bigr]}_{\text{masked likelihood}},
    \label{eq:bayesian}
\end{equation}
where $\mathcal P_{\text{phys}}$ is the physics surrogate restricts $\mathbf V$ to fields that satisfy the physics constraint.  $p_\theta(\mathbf V)$ is the mask-aware prior learned from historical data. The masked likelihood incorporates the measurements only at the observed entries specified by $\mathbf M$.
Together, these terms define the distribution of all speed fields that are physically reasonable, consistent with historical data, and match the available observations.
\begin{remark}
Bayes' rule factorizes the posterior as
\begin{equation}
    p(\mathbf V\mid \mathbf Y) 
    \propto p(\mathbf V) \cdot p(\mathbf Y\mid \mathbf V).
\end{equation}
Because the noise acts only on visible pixels, the likelihood becomes
\begin{equation}
    p(\mathbf Y\mid \mathbf V)
    =\prod_{(s,t):\,M_{s,t}=1}
      \frac{1}{\sqrt{2\pi}\,\sigma}\,
      \exp\!\left[
            -\frac{(Y_{s,t}-V_{s,t})^{2}}{2\sigma^{2}}
          \right],
\end{equation}
which is exactly the exponential term in Eq.~\eqref{eq:bayesian}. The likelihood depends only on the measurements on the support of $\mathbf M$ and leaves unobserved entries unconstrained. The prior $p_\theta(\mathbf V)$ introduces data-driven regularization, and the indicator $\mathbf 1[\mathbf V\in\mathcal P_{\text{phys}}]$ restricts the support to speed fields that satisfy the surrogate physics constraints.
\end{remark}

To sample from Eq.~\eqref{eq:bayesian}, we introduce a mask-aware diffusion prior training strategy in Section~\ref{sec:ambient} and a physics-guided sampling strategy in Section~\ref{sec:sampler} that alternates reverse diffusion, observation projection, and physics projection.

\subsection{Mask-Aware Diffusion Prior}
\label{sec:ambient}
To learn the data-driven prior \(p_\theta(\mathbf V)\) that appears in the Bayesian posterior (Eq.~\ref{eq:bayesian}), we must train a generative model directly from the same sparsely observed speed fields available at inference. However, classical denoising-diffusion probabilistic models (DDPMs, \cite{ho2020denoising}) require fully observed training data. This condition is rarely satisfied in highway traffic state estimation, where each speed field $\mathbf{V} \in \mathbb{R}^{S \times T}$ is only partially revealed through a binary mask $\mathbf{M}$. If such incomplete observations are treated as inputs, the learned prior may fail to capture the true variability in unobserved regions. To overcome this mismatch, we employ an Ambient Diffusion \citep{daras2023ambient} strategy, which integrates the masking process directly into the data distribution.

Specifically, during training we draw $\tilde{\mathbf M}\sim \pi(\mathbf M_0)$ from the same law expected at inference, combining fixed loop detector positions with randomly sampled probe trajectories. Here, $\pi$ denotes the empirical distribution of masks observed in the raw data. The partially observed input is then constructed as 
\begin{equation}
    \mathbf Y_0=\tilde{\mathbf M}\odot (\mathbf V + \boldsymbol\varepsilon).
\end{equation}

The denoiser is conditioned on the current mask $\tilde{\mathbf M}$, so its prediction uses the observed values while recognizing which entries are missing. Under Assumptions~\ref{asm:indep} and~\ref{asm:mar}, the masked denoising objective corresponds to learning the conditional score on the unobserved entries given the observed ones, even though a full $\mathbf V$ is never available. This allows the sensing sparsity to be incorporated directly into the learned prior without relying on external preprocessing or imputation.

\subsubsection{Mask-selection strategies}
Observation masks are heterogeneous: some pixels (e.g., near fixed detectors) are almost always visible, while others remain persistently hidden. To account for this, we consider two complementary training strategies summarized in Table~\ref{tab:mask-strategies}.

\begin{figure}[width=.99\linewidth,cols=4,pos=t]
    \centering
    \includegraphics[width=0.99\linewidth,
        trim=15mm 92mm 15mm 0mm,
        clip]{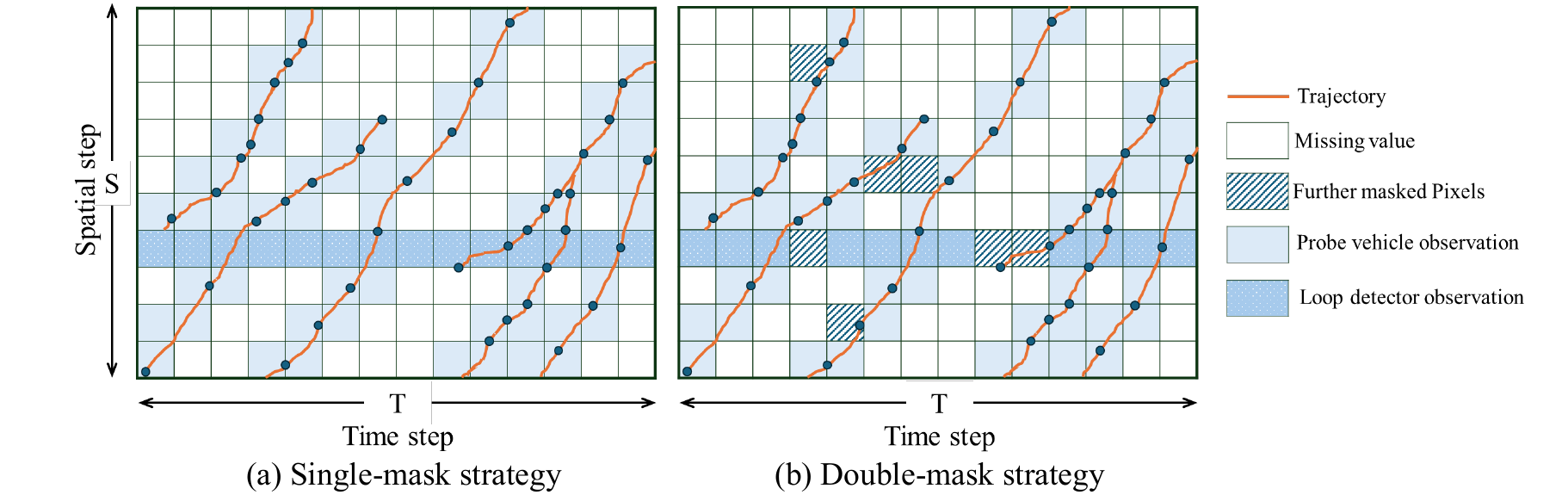}
    \caption{Illustration of two mask strategy. Left: original observed data under the single-mask strategy, reflecting actual sparse sensor coverage. Right: the same frame with further masked pixels, simulating the observation pattern under the double-mask strategy.}
    \label{fig:two mask strategy}
\end{figure}

In the single mask strategy, the true sensor mask $\mathbf M$ is left untouched; the network is therefore penalized only on truly known pixels. This works well when probe vehicle coverage is dense and roughly random. Under such conditions every spatial-temporal position is observed at least occasionally.
In the double-mask  strategy, when the overall visibility ratio
$\rho=\lVert\mathbf M\rVert_{0}/(ST)$ is high, we draw the auxiliary mask $\mathbf B$ from the empirical mask distribution $\pi(\mathbf M)$ captured in the raw data.  When $\rho$ is low, we instead sample $\mathbf B$ from a Bernoulli distribution with parameter $p_{\text{extra}}$, so that every coordinate is occasionally masked and therefore contributes to the loss.
In practice we use \(p_{\text{extra}}=0.05\), which improves interpolation near fixed loop locations while incurring negligible additional variance, see Figure~\ref{fig:two mask strategy}.
\begin{table}[width=.99\linewidth,cols=4,pos=b]
    \centering
    \caption{Mask-selection strategies for training the diffusion prior.}
    \begin{tabular}{@{\hspace{4pt}}lccp{8cm}@{\hspace{4pt}}}
    \toprule
    \textbf{Strategy}  & \textbf{Pixels kept} & \textbf{Regions contributing to loss} & \textbf{When to prefer}\\
    \midrule
    Single mask  &$\mathbf M_{s,t}=1$& \(\mathbf M\) &
    All pixels have non-zero probability of being observed.\\[2pt]
    Double mask  &$\mathbf M_{s,t}=1$& \(\tilde{\mathbf M}=\mathbf B\odot\mathbf M\) &
    Some pixels are always invisible (e.g.\ fixed loop detectors)\\
    \bottomrule
    \label{tab:mask-strategies}
    \end{tabular}
\end{table}

\subsubsection{Mask-aware training process}
To construct a mask-aware prior, we explicitly mimic the sensing sparsity
during training.  For each speed field \(\mathbf V\) we first
sample a training mask $\tilde{\mathbf M}$ (according to the single- or
double-mask strategy) and build
\begin{equation}
\mathbf Y_0=\mathcal C_\rho(\mathbf V)=\tilde{\mathbf M}_{s,t}\odot(\mathbf V + \boldsymbol\varepsilon),
\end{equation}
where $\tilde{\mathbf M}_{s,t}=1$ marks an observed pixel, $\mathcal C_\rho$ denotes the masking operator with the ratio $\rho$. Next we corrupt \(\mathbf Y_0\) with Gaussian noise first before asking the network to denoise it.

\noindent \textbf{Step 1 - Forward corruption (variance-preserving SDE).}
We gradually add Gaussian noise to \(\mathbf Y_0\) using the
variance-preserving SDE
\begin{equation}
    \mathrm d\mathbf Y_t = -\frac12\,\beta(t)\,\mathbf Y_t\,\mathrm dt + \sqrt{\beta(t)}\,\mathrm d\mathbf W_t , \qquad \beta(t)=\beta_0+(\beta_1-\beta_0)t,
\end{equation}
where \(\mathbf W_t\) is a standard Wiener process. Then the closed-form solution for \(t\in[0,1]\) is
\begin{equation}
    \mathbf Y_t= \sqrt{\alpha(t)}\,\mathbf Y_0+ \sqrt{1-\alpha(t)}\,\boldsymbol\eta ,\qquad\boldsymbol\eta\sim\mathcal N(\mathbf 0,\mathbf I),\;\alpha(t)=\exp\!\Bigl(-\!\int_0^t\!\beta(s)\,\mathrm ds\Bigr).
\end{equation}
Thus, as $t$ increases, $\alpha(t)$ decreases; and the further the diffusion proceeds, the closer \(\mathbf Y_t\) becomes to pure noise.

\noindent\textbf{Step 2 - Masked score matching (noise prediction).}
Given $(\mathbf Y_t,t,\mathbf M)$, the denoiser $h_\theta(\cdot)$ is trained to predict the injected noise $\boldsymbol\eta$, but only where we have data:
\begin{equation}
        \mathcal L(\theta)=\mathbb E\!\bigl[\lambda(t)\,\|\mathbf M\odot(h_\theta(\mathbf Y_t,t,\mathbf M)-\boldsymbol\eta)\|_F^{2}\bigr],\qquad
        \lambda(t)=\alpha(t)/(1-\alpha(t)).
        \label{eq:mask-noise-loss}
\end{equation}
Here \(\lambda(t)\) re-weights time steps according to the signal-to-noise ratio, and the Frobenius norm \(\|\cdot\|_F\) sums squared errors over \(\mathbf M_{s,t}=1\).
Under the standard noise-prediction parameterization \citep{daras2023ambient}, the conditional score for unobserved pixels ($\mathbf M_{s,t}=0$) is approximated by
\begin{equation}
    \nabla_{\mathbf Y_t}\!\log p_t(\mathbf Y_t)
    \;\approx\;
    -\frac{1}{\sigma_t}\,h_\theta(\mathbf Y_t,t,\mathbf M),
    \label{eq:score-approx}
\end{equation}
and Assumption~\ref{asm:mar} ensures that the loss in Eq.~\eqref{eq:mask-noise-loss} matches the corresponding masked score-matching objective. We therefore view $h_\theta$ as a data-driven approximation of the conditional score under the same masking law that will be encountered at inference time.

Overall, the procedure creates a stochastic corruption-restoration process: we corrupt a sparsely observed speed field with noise (forward SDE) and train the network to reverse that corruption (masked score matching). Unlike imputation-based pre-processing, which risks propagating bias from heuristic gap-filling, the proposed strategy trains the prior directly from incomplete observations. 
As a result, the diffusion prior $p_\theta(\mathbf V)$ captures the spatiotemporal dynamics of traffic speeds without requiring fully observed training data.

\subsection{Physics-Guided Posterior Sampler}
\label{sec:sampler}
While the learned diffusion prior $p_\theta$ effectively captures the distribution of highway speed fields, it is agnostic to two essential constraints: (i) strict fidelity to partial observations $(\mathbf Y,\mathbf M)$ and (ii) consistency with macroscopic traffic-flow principles. Naively conditioning on observations through noise replacement or direct imputation often leads to sharp discontinuities and physically implausible structures in the reconstructed regions, particularly under severe sparsity.
Conversely, exact enforcement of conservation laws is infeasible in the present setting, where only normalized speeds are observed and neither density nor flow can be reliably recovered.
To address this gap, we introduce a surrogate physics posterior that integrates three complementary elements: 
a data-driven prior, exact observation fidelity, and a soft projection toward physically reasonable patterns. In practice, we implement this idea through a physics-guided posterior sampler. We view the resulting sampling strategy as a heuristic algorithm for approximately exploring the surrogate posterior in Eq.~\eqref{eq:bayesian}, rather than as an exact Markov chain Monte Carlo method, and we do not claim theoretical convergence to the target distribution.

\subsubsection{Surrogate physics posterior}
Let $p_\theta(\mathbf V)$ denote the mask-aware diffusion prior. Our target distribution for inference is the surrogate physics posterior, see Eq.~\ref{eq:bayesian}. Let $\mathcal P_{\text{phys}}\subset[0,1]^{S\times T}$ denotes a surrogate physics set, implicitly defined by a mask-invariant projection operator
\begin{equation}
    \mathcal P_{\text{phys}}:\;
        \underbrace{[0,1]^{S\times T}}_{\text{current speed field }\mathbf V}
        \times
        \underbrace{\{0,1\}^{S\times T}}_{\text{mask }\mathbf M}
        \;\xrightarrow{ P_{\text{phys}}} \;
        \underbrace{[0,1]^{S\times T}}_{\text{updated field}}.
\end{equation}
Here, we approximate it through an outer constrained reverse process that alternates reverse-diffusion updates with observation and physics projections while incorporating a physics-guided projector.

\subsubsection{Physics-guided posterior sampler}
At each reverse-diffusion iteration, the sampler alternates three operations:
(i) \textbf{Denoise:} Use the denoiser to draw the current sample one step closer to the learned prior, 
(ii) \textbf{Observation projection:} Replace every visible pixel with its exact measurement to preserve data fidelity, 
and (iii) \textbf{Physics projection:} Project the missing pixels onto the low-frequency manifold $\mathcal P_{\text{phys}}$ defined by the traffic surrogate.
An optional forward jump re-injects a small amount of forward noise, then resumes the reverse process. 

In our setting, we draw $\mathbf Z_T\sim\mathcal N(\mathbf 0,\mathbf I)$ with a linear $\beta$-schedule of length $T=500$.
For $t=T{-}1,\ldots,0$, perform:
\begin{align}
\textit{Reverse diffusion:}&\quad
\tilde{\mathbf Z}_{t}=g_{\theta}(\mathbf Z_{t+1},t+1),
\label{eq:sampler-rev}\\[4pt]
\textit{Observation projection:}&\quad
\hat{\mathbf Z}_{t}=P_{\text{obs}}(\tilde{\mathbf Z}_{t})
=\mathbf Y\odot\mathbf M+\tilde{\mathbf Z}_{t}\odot(1-\mathbf M),
\label{eq:sampler-obs}\\[4pt]
\textit{Physics projection:}&\quad
\bar{\mathbf Z}_{t}=P_{\text{phys}}(\hat{\mathbf Z}_{t},\mathbf M),
\label{eq:sampler-phys}\\[4pt]
\textit{Optional forward jump:}&\quad
\mathbf Z_{t}=f_{j}(\bar{\mathbf Z}_{t},t)
=\sqrt{\frac{\alpha(t)}{\alpha(t+j)}}\,\bar{\mathbf Z}_{t}
+\sqrt{1-\frac{\alpha(t)}{\alpha(t+j)}}\,\boldsymbol\epsilon.
\label{eq:sampler-fwd}
\end{align}

where $\mathbf Z_t$ denotes the sampler state at reverse step $t$, $g_{\theta}$ denotes the reverse diffusion kernel, and $f_j$ denotes the forward jump operator.
The observation projection (\ref{eq:sampler-obs}) ensures exact data fidelity on $\mathbf M=1$, while the physics projector (\ref{eq:sampler-phys}) acts only on the missing subspace, enforcing mask-invariance:
\begin{equation}
    P_{\text{phys}}(\mathbf{V},\mathbf M)\odot\mathbf M = \mathbf V\odot\mathbf M.
\end{equation}
Repeating this constrained loop yields $\{\mathbf Z_0^{(n)}\}_{n=1}^N$, an approximate ensemble of posterior samples for uncertainty quantification.

\subsubsection{Physics-guided projector.}
The current instantiation of $P_{\text{phys}}$ employs an Adaptive Anisotropic Smoothing (AAS) operator, which extends the Generalized Adaptive Smoothing Method (GASM)~\citep{treiber2011reconstructing} into a mask-aware and diffusion compatible form. 
Unlike full conservation enforcement, which is infeasible under speed-only observations, AAS acts as a local regularizer that promotes physically plausible structures in unobserved regions while preserving exact fidelity on measured pixels, see Algorithm~\ref{alg:aas-projector}.
Its design is guided by three principles:
\begin{enumerate}
    \item \textbf{Characteristic coherence.} Anisotropic smoothing aligns kernel support with admissible forward/backward slopes $c\in\{c_{\text{free}},c_{\text{cong}}\}$ (after conversion by $\Delta t/\Delta x$), approximating propagation along traffic waves instead of isotropic blurring.
    \item \textbf{Soft regime transition:} Instead of a hard free-flow or congestion split, AAS introduces a differentiable gate $p_{\text{free}}=\frac12(1+\tanh((V_{s,t}-v_{\text{thr}})/\beta))$ that blends the free-flow and congested kernels.
    \item \textbf{Bounded first-order transport residual}: A residual term 
    $R_{s,t} = (V_{s,t}-V_{s,t-1}) + c_{s,t}(V_{s,t}-V_{s-1,t})$
    penalizes local violations of the first-order kinematic wave relation 
    $\partial_t V + c\,\partial_x V \approx 0$, providing a lightweight consistency proxy without requiring explicit density or flow estimates.
\end{enumerate}
The resulting update is applied only on $(1-\mathbf M)$:
\begin{equation}
    V^{(1)}=V-\alpha_{\text{smooth}}(V-V_s)\odot(1-\mathbf M),\qquad
    V^{(2)}=V^{(1)}-\alpha_{\text{char}}R\odot(1-\mathbf M),
    \label{eq:relax}
\end{equation}
where $V_s$ is the blended anisotropic surrogate field and $\alpha_{\text{smooth}}\gg\alpha_{\text{char}}$ ensures numerical stability. 
A final clipping step projects the result onto $[0,1]$. 
By construction, the update is mask-invariant and acts only on the missing subspace, and the step sizes $(\alpha_{\text{smooth}},\alpha_{\text{char}})$ are chosen so that the projector remains numerically stable under iterative application within the diffusion loop.

The physics-guided posterior sampler combines the generative prior with a physics projector that enforces basic traffic-flow structure on the missing region. A central design feature is modularity: any projector that is (i) mask-invariant, (ii) non-expansive on the missing subspace, and (iii) computationally local can replace or augment AAS without retraining the prior $p_\theta$. The same interface also allows multiple physics modules to be applied in sequence. If additional variables such as density, flow, or ramp counts become available, projectors based on conservation relations or fundamental-diagram consistency (e.g., $P_{\text{cons}}$, $P_{\text{FD}}$) can be inserted into the sampling loop. In the current speed-only setting we use only $P_{\text{AAS}}$, but the framework accommodates richer physics when appropriate data are available.
\begin{algorithm}[t]
\caption{\textbf{AAS Projector: Physics-guided Mask-invariant Projection}}
\label{alg:aas-projector}
\begin{algorithmic}[1]
\Require Speed field $V \in [0,1]^{S\times T}$; mask $\mathbf M \in \{0,1\}^{S\times T}$; 
physical parameters $(v_{\text{thr}}, \beta, c^{\text{phys}}_{\text{free}}, c^{\text{phys}}_{\text{cong}})$; 
smoothing weights $(\alpha_{\text{smooth}}, \alpha_{\text{char}})$; 
kernel stds $(\sigma_{par}, \sigma_t)$.
\Ensure Updated field $P_{\text{phys}}(V,\mathbf M)$.

\State Compute soft regime gate:
\[
p_{\text{free}} \gets 0.5\bigl(1+\tanh((V - v_{\text{thr}})/\beta)\bigr)
\]
\State Convert characteristic speeds to grid units:
\(
c_{\text{free}} \gets c^{\text{phys}}_{\text{free}}\Delta t/\Delta x;\;
c_{\text{cong}} \gets c^{\text{phys}}_{\text{cong}}\Delta t/\Delta x
\)
\State Build two anisotropic Gaussian kernels $K_{c_{\text{free}}}$, $K_{c_{\text{cong}}}$ 
aligned with slopes $c_{\text{free}}, c_{\text{cong}}$ (normalized).
\State Apply separable convolution:
\(
V_f \gets K_{c_{\text{free}}}*V,\; V_c \gets K_{c_{\text{cong}}}*V
\)
\State Compute blended surrogate:
\(
V_s \gets p_{\text{free}}\odot V_f + (1-p_{\text{free}})\odot V_c
\)
\State \textbf{Step 1 (smooth relax):}
\(
V^{(1)} \gets V - \alpha_{\text{smooth}}(V - V_s)\odot(1-\mathbf M)
\)
\If{transport correction enabled}
    \State Compute local speed $c_{\text{loc}}\gets p_{\text{free}}c_{\text{free}}+(1-p_{\text{free}})c_{\text{cong}}$
    \State Compute residual:
    \(
    R \gets (V^{(1)}_{s,t}-V^{(1)}_{s,t-1}) + c_{\text{loc}}(V^{(1)}_{s,t}-V^{(1)}_{s-1,t})
    \)
    \State \textbf{Step 2 (residual update):}
    \(
    V^{(2)} \gets V^{(1)} - \alpha_{\text{char}}R\odot(1-\mathbf M)
    \)
\Else
    \State $V^{(2)} \gets V^{(1)}$
\EndIf
\State Clip to valid range:
\(
V^{(2)} \gets \text{clip}(V^{(2)}, 0, 1)
\)
\State \Return $P_{\text{phys}}(V,\mathbf M)=V^{(2)}$
\end{algorithmic}
\end{algorithm}

\section{Experimental Evaluation} 
\label{sec:experiments}
We conduct a comprehensive empirical evaluation to assess the effectiveness of \textbf{PMA-Diffusion}.
The experiments are structured around two central research questions,  each aligned with one stage in the PMA-Diffusion framework:
\begin{description}
  \item[RQ1 -- Prior learning:] How well a diffusion prior can be learned on incomplete speed fields?
  \item[RQ2 -- Posterior sampling:] Given a fixed prior, how much the sampling stage can recover under varying sensing sparsity?
\end{description}
PMA-Diffusion consists of two main components: (i) a mask-aware diffusion prior trained directly from sparsely observed speed fields, and (ii) a physics-guided posterior sampler that alternates reverse-diffusion updates, observation projection, and physics-guided projection. \textbf{RQ1} evaluates whether a diffusion prior trained under sparse supervision (Single-mask and Double-mask training strategies) can approach the performance of a model trained on fully observed speed fields. \textbf{RQ2} then examines how much the proposed posterior sampler can further improve reconstruction accuracy by enforcing data fidelity and basic kinematic-wave structure under different sensing sparsity levels.
\begin{figure}[width=.99\linewidth,cols=4,pos=t]
    \centering
    \includegraphics[width=0.7\linewidth]{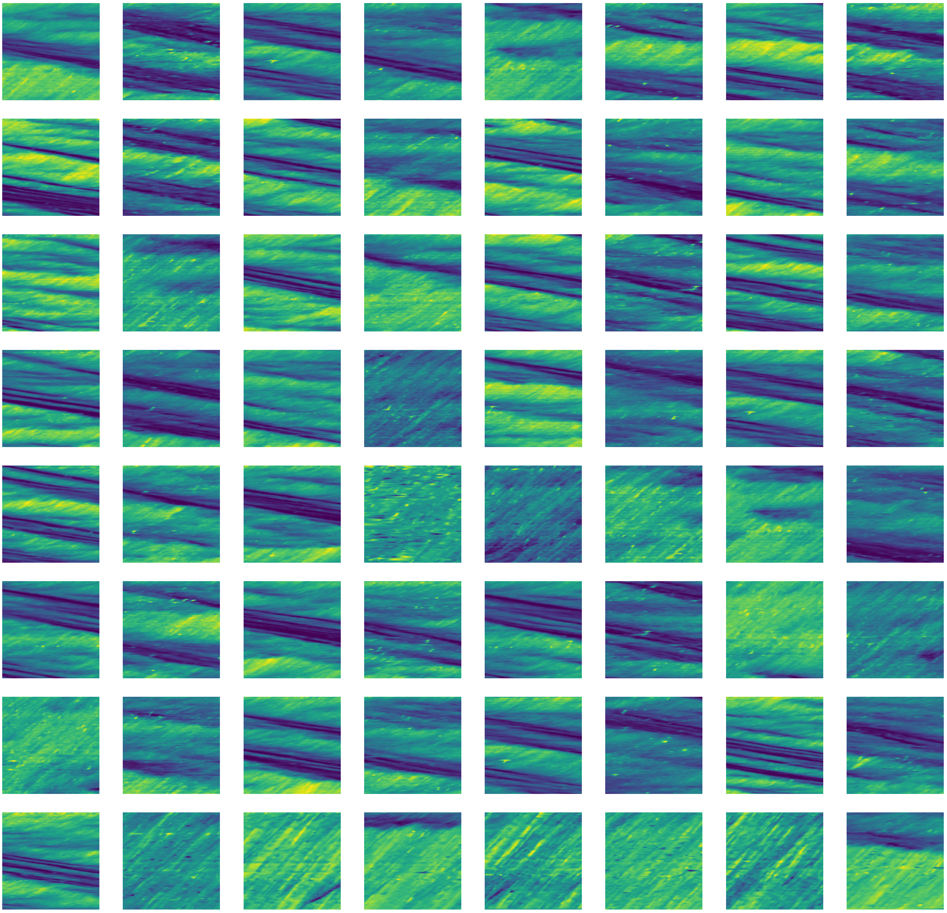}
    \caption{Random speed field samples from I-24 MOTION dataset. These samples illustrate the spatial-temporal structure (propagating waves, lane heterogeneity) that the model is required to learn.}
    \label{fig:randomSample}
\end{figure}

\subsection{Dataset, Pre-processing, Experimental Set-up}

The study corridor corresponds to the Tennessee Department of Transportation's I-24 Mobility Technology Interstate Observation Network (I-24 MOTION; \cite{gloudemans202324}) data. I-24 MOTION data are captured by 276 pole-mounted high-resolution traffic cameras that provide seamless coverage of approximately 4.2 miles along the I-24 highway corridor near Nashville, TN. 
In our experiment, we convert vehicle trajectories into lane-level speed fields on a fixed Eulerian grid with $S=200$ spatial pixels ($\Delta x\!=\!100$ft), four lanes, and $T=200$ time slices ($\Delta t\!=\!1$ s), so that each sample spans approximately 3.8 miles over 200 seconds. Figure~\ref{fig:randomSample} shows randomly selected samples from the dataset. The dataset is split into 70\% / 10\% / 20\% for training, validation, and testing, respectively. and all quantitative evaluations are conducted on test samples.

The I-24 MOTION offers two advantages for this study. First, its near-complete spatial and temporal coverage yields a "ground-truth" speed field, which allows us to mask the data synthetically and evaluate reconstruction error at every hidden pixel. Second, it includes dense and clean vehicle trajectories that capture real traffic movements without the gaps and noise found in typical probe vehicle data. By sub-sampling I-24 MOTION to mimic realistic detector layouts and probe vehicle trajectories, we can test PMA-Diffusion under controlled sparsity levels while still comparing the results against a fully observed baseline.

\subsubsection{Observation models} 
\begin{figure}[width=.99\linewidth,cols=4,pos=t]
  \centering
  \captionsetup{justification=centering}  
  \begin{subfigure}[t]{0.24\textwidth}
    \centering
    \includegraphics[width=\linewidth]{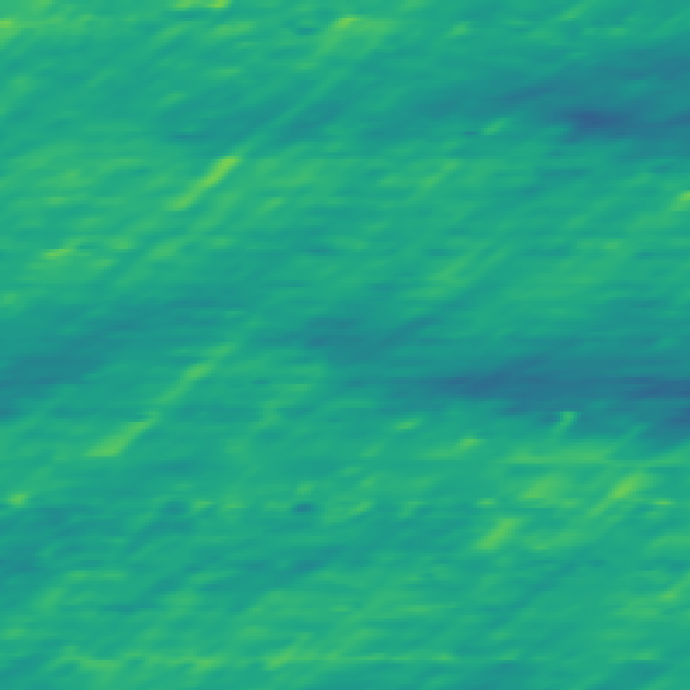}
    \caption{Ground-truth data}
  \end{subfigure}
  \hfill
  \begin{subfigure}[t]{0.24\textwidth}
    \centering
    \includegraphics[width=\linewidth]{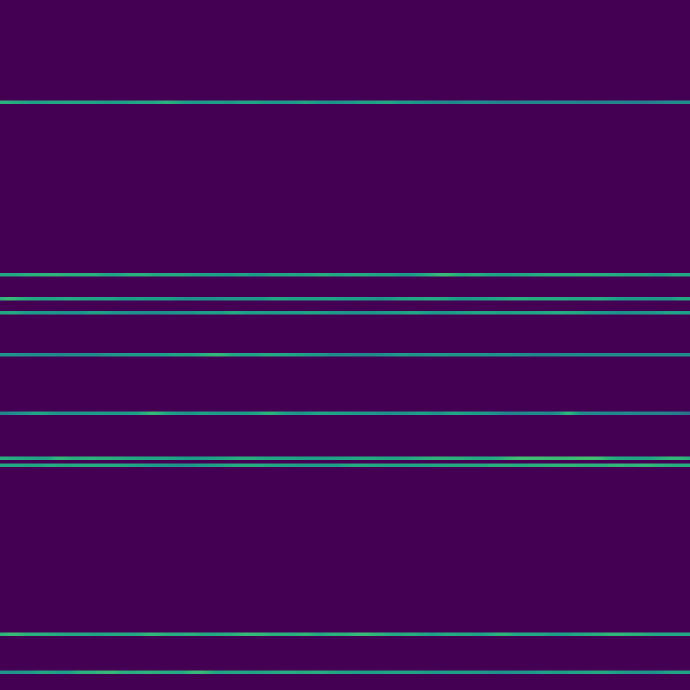}
    \caption{Loop detector}
  \end{subfigure}
  \hfill
  \begin{subfigure}[t]{0.24\textwidth}
    \centering
    \includegraphics[width=\linewidth]{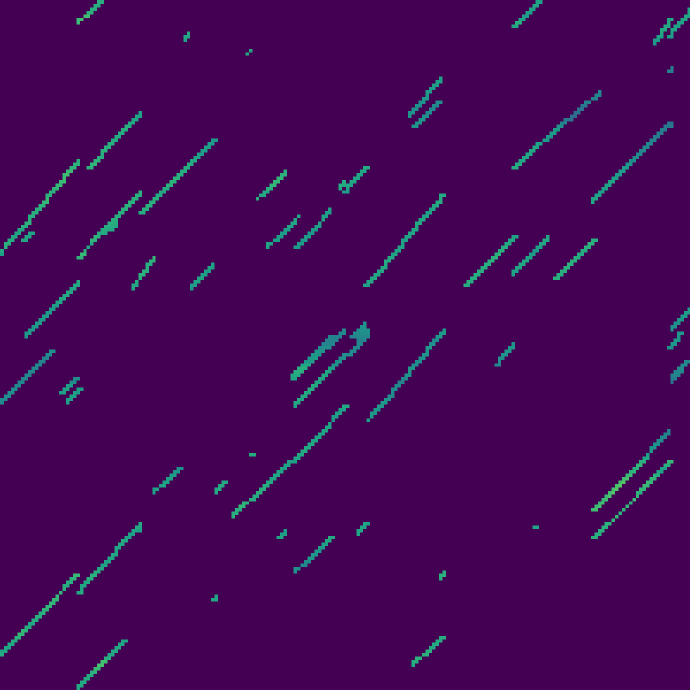}
    \caption{Probe vehicle}
  \end{subfigure}
  \hfill
  \begin{subfigure}[t]{0.24\textwidth}
    \centering
    \includegraphics[width=\linewidth]{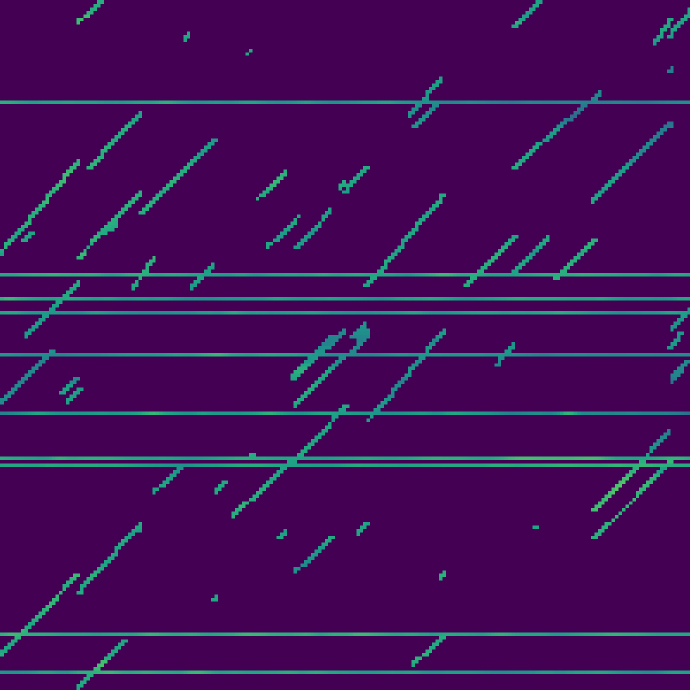}
    \caption{Total observed}
  \end{subfigure}
  \caption{Observation Patterns from Loop detectors and Probe vehicles. Dark areas denote unobserved pixels; colored entries denote observed speeds.}
  \label{fig:exp_obs}
\end{figure}

We collect synthetic observation ($\mathbf{Y}$) and mask ($\mathbf{M}$) pairs based on the ground-truth speed field ($\mathbf{V}$) to simulate the real-world scenarios. Our masks have two components:
\begin{itemize}
    \item \textbf{Loop detectors mask:} In practice, the Loop detectors are fixed in a specific location. We generate a fixed Loop detector mask $\mathbf M^{(l)}$ and mask $5\%, 15\%, 25\%$ of the row data as observed data $\mathbf M^{(l)}\odot\mathbf V$.
    \item \textbf{Probe vehicles mask:} To simulate probe vehicle data, we use real I-24 MOTION trajectories to build observation masks. We randomly select a subset of vehicle IDs from the preprocessed data and reconstruct their trajectories using position, timestamp, and speed information. In the experiments, we evaluate four probe density levels, corresponding to $\lambda = 0, 5, 15,$ and $25$, respectively.
\end{itemize}
Building on Assumption~\ref{asm:latency} and Assumption~\ref{asm:drift}, we resample the probe vehicle trajectories directly onto the Eulerian grid. This ensures that the probe vehicle observations align with the grid and removes small misalignment effects. Figure~\ref{fig:exp_obs} shows the ground-truth speed field and the three corresponding observation masks: (b) the loop detector mask, (c) the probe vehicle mask obtained from resampled trajectories, and (d) the total observed mask formed by their union.

\subsection{Evaluation Metrics}
We evaluate reconstruction quality only on the unobserved pixels, because the observed pixels are directly replaced by measurements during sampling and therefore do not indicate how well the model can recover missing states. Meanwhile, our goal is not to assess point-wise prediction accuracy, but to measure whether the generated speed fields are reasonable, physically plausible, and consistent with macroscopic traffic patterns. Following \citet{choi2025gentle}, we use three complementary metrics.

\begin{enumerate}
  \item \textbf{Masked-MSE (2$\times$2).} We first apply a $2\times2$ average pooling to both the reconstruction and the ground truth, and then compute the mean squared error on the unobserved pixels. This metric measures the accuracy of the low-frequency component of the speed field, which is most relevant for traffic operations, such as identifying the onset and extent of congestion, and for computing link travel times. The pooling step reduces sensitivity to very local differences (e.g., a one-pixel shift of a wave front) that are less important at the macroscopic level.
  \item \textbf{Sobel-MSE.} We apply a Sobel operator to extract spatial and temporal gradients of the speed field, and then compute the mean squared error between the edge maps of the reconstruction and the ground truth on the unobserved pixels. This metric emphasizes the reconstruction of high-frequency structures, in particular the location and sharpness of stop-and-go waves and shock fronts, which are central to many traffic-flow analyses.
  \item \textbf{LPIPS.} We report the Learned Perceptual Image Patch Similarity (LPIPS) score with an AlexNet backbone \citep{zhang2018unreasonable} on the unobserved pixels. LPIPS compares feature representations in a pretrained network and serves as a perceptual measure of similarity between two spatiotemporal speed fields. It captures differences in pattern shape and texture that may not be fully reflected in pixel-wise errors but matter for visual inspection and qualitative traffic pattern analysis.
\end{enumerate}

\subsection{Models}
The three training and sampling strategies introduced below are structured specifically to isolate the contributions of the prior and the sampling mechanism with respect to RQ1 and RQ2, respectively.
\subsubsection{Training strategy}
\label{exp:training}
We evaluate three training strategies designed to isolate the effects of supervision.
\begin{itemize}
    \item  \textbf{Full-obs} The network trained by fully latent speed field data $\mathbf V$; the value of \textbf{Full-obs} represents the upper bound, showing how well the DDPM architecture alone can fit fully observed data.
    \item \textbf{Single-mask} Each training sample is corrupted by its own visibility mask, exactly mirroring realistic loop detectors + probe vehicles coverage. The model therefore learns only from genuinely observed pixels.
    \item \textbf{Double-mask} On top of the true visibility map we apply an additional mask (see Table \ref{tab:mask-strategies}). Hiding even "always visible" rows encourages the network to reason about persistently unseen locations and improves robustness when some pixels are never measured in practice.
\end{itemize}
All three variants share the same UNet backbone, which is the most widely used backbone neural network structure for diffusion models \citep{ronneberger2015u, ho2020denoising}.
Specifically, our UNet backbone is configured with one input channel, 48 base feature channels, four resolution levels with channel multipliers of $ (1, 2, 4, 8)$ and 4-head linear attention in the bottleneck.
We use a linear $\beta$-schedule with $T{=}500$ steps from $\beta_0=10^{-4}$ to $\beta_T=2{\times}10^{-2}$ and optimize the Huber loss ($\delta=1$), which is numerically more stable than $\ell_2$.
We train for 50 epochs using Adam (lr $=5\!\times\!10^{-4}$, batch size $=64$). The model training time is around 300 minutes of GPU time. All experiments are conducted on a shared GPU server equipped with two AMD EPYC Milan 7643 CPUs, four NVIDIA RTX 6000 Ada Generation GPUs with 48\,GB GDDR6 memory, 1\,TB system memory, and NVMe storage. In this study we only use a single NVIDIA RTX 6000 Ada GPU for training and inference.

\subsubsection{Sampling strategy}
\label{sec:inference}
Using the diffusion prior from the training stage, we compare two main diffusion based sampling strategies and one physics-only baseline. 
\begin{enumerate}
\item \textbf{AAS-only baseline.}
    We start from the partially observed field $\mathbf Y$ and apply the AAS projector for a number of iterations. 
    At each iteration, the observed pixels (mask $M=1$) are kept unchanged, and the projector updates only the missing pixels ($M=0$) by anisotropic smoothing and a small first-order correction. 
    This baseline does not use the diffusion model.
\item \textbf{RePaint.}
    We follow the RePaint inpainting strategy: the sampler alternates reverse-diffusion steps with occasional small forward noising steps, while repeatedly resetting the observed pixels to their measurements to better satisfy the observation mask.
\item \textbf{PMA-Diffusion.}
    We start from the same inpainting strategy and, after enforcing the observations, apply the AAS projector on the unobserved pixels, so that the missing region is guided toward simple first-order traffic patterns without changing the observed data.
\end{enumerate}
The $\beta$-schedule and timestep ($T{=}500$) are shared with training stage. RePaint uses the same U-Net to predict $\epsilon_\theta$; the physics projector is parameter-free at test time once the hyper-parameters are fixed. 

\subsection{Experimental Results}
\label{sec:exp_results}

\begin{table}[width=.99\linewidth,cols=4,pos=b]
\centering
\scriptsize
\setlength{\tabcolsep}{4pt}
\caption{Quantitative comparison of three training modes (M1: Full-obs, M2: Single-mask, M3: Double-mask) on $200{\times}200$ grids under different row coverage levels (Row) and probe intensities ($\lambda$). For each setting we report LPIPS, Masked-MSE (2$\times$2), and Sobel-MSE for three reconstruction methods (AAS-only, RePaint, PMA-Diffusion). All metrics are computed on unobserved pixels; lower values indicate better reconstruction.}
\label{tab:full_result}
\begin{tabular}{llc rrr rrr rrr}
\toprule
\multirow{2}{*}{Mode} & \multirow{2}{*}{Row} & \multirow{2}{*}{$\lambda$} &
\multicolumn{3}{c}{LPIPS} &
\multicolumn{3}{c}{Masked-MSE (2$\times$2)} &
\multicolumn{3}{c}{Sobel-MSE} \\
\cmidrule(lr){4-6}\cmidrule(lr){7-9}\cmidrule(l){10-12}
& & & AAS-only & RePaint & PMA & AAS-only & RePaint & PMA & AAS-only & RePaint & PMA \\
\midrule
\multirow{12}{*}{M1}
& \multirow{4}{*}{5\%}
& 0  & 0.2662 & \textbf{0.1617} & 0.1768 & 0.1463 & 0.0172 & \textbf{0.0167} & 0.0028 & 0.0013 & \textbf{0.0009} \\
& & 5  & 0.2466 & 0.1491 & \textbf{0.1357} & 0.1356 & 0.0155 & \textbf{0.0113} & 0.0030 & 0.0014 & \textbf{0.0008} \\
& & 15 & 0.2275 & 0.1345 & \textbf{0.0981} & 0.1167 & 0.0152 & \textbf{0.0064} & 0.0033 & 0.0017 & \textbf{0.0008} \\
& & 25 & 0.2117 & 0.1299 & \textbf{0.0800} & 0.1009 & 0.0158 & \textbf{0.0043} & 0.0034 & 0.0019 & \textbf{0.0008} \\
\cmidrule(lr){2-12}
& \multirow{4}{*}{15\%}
& 0  & 0.1187 & 0.0823 & \textbf{0.0381} & 0.0523 & 0.0195 & \textbf{0.0018} & 0.0043 & 0.0038 & \textbf{0.0006} \\
& & 5  & 0.1179 & 0.0854 & \textbf{0.0353} & 0.0487 & 0.0185 & \textbf{0.0017} & 0.0042 & 0.0037 & \textbf{0.0006} \\
& & 15 & 0.1169 & 0.0908 & \textbf{0.0311} & 0.0421 & 0.0165 & \textbf{0.0016} & 0.0039 & 0.0035 & \textbf{0.0006} \\
& & 25 & 0.1140 & 0.0936 & \textbf{0.0294} & 0.0367 & 0.0146 & \textbf{0.0014} & 0.0036 & 0.0033 & \textbf{0.0006} \\
\cmidrule(lr){2-12}
& \multirow{4}{*}{25\%}
& 0  & 0.0821 & 0.0657 & \textbf{0.0322} & 0.0416 & 0.0104 & \textbf{0.0018} & 0.0032 & 0.0022 & \textbf{0.0006} \\
& & 5  & 0.0848 & 0.0701 & \textbf{0.0300} & 0.0387 & 0.0099 & \textbf{0.0017} & 0.0031 & 0.0022 & \textbf{0.0006} \\
& & 15 & 0.0843 & 0.0769 & \textbf{0.0266} & 0.0336 & 0.0089 & \textbf{0.0015} & 0.0029 & 0.0022 & \textbf{0.0006} \\
& & 25 & 0.0836 & 0.0828 & \textbf{0.0253} & 0.0294 & 0.0083 & \textbf{0.0014} & 0.0027 & 0.0022 & \textbf{0.0006} \\
\midrule
\multirow{12}{*}{M2}
& \multirow{4}{*}{5\%}
& 0  & 0.2662 & 0.8455 & \textbf{0.1851} & 0.1463 & 7.1959 & \textbf{0.0364} & 0.0028 & 0.3112 & \textbf{0.0008} \\
& & 5  & 0.2466 & 0.3540 & \textbf{0.1406} & 0.1356 & 0.1727 & \textbf{0.0133} & 0.0030 & 0.0116 & \textbf{0.0008} \\
& & 15 & 0.2275 & 0.4048 & \textbf{0.0981} & 0.1167 & 0.1105 & \textbf{0.0052} & 0.0033 & 0.0058 & \textbf{0.0008} \\
& & 25 & 0.2117 & 0.1966 & \textbf{0.0821} & 0.1009 & 0.0199 & \textbf{0.0042} & 0.0034 & 0.0029 & \textbf{0.0008} \\
\cmidrule(lr){2-12}
& \multirow{4}{*}{15\%}
& 0  & 0.1187 & 1.2911 & \textbf{0.0367} & 0.0523 & 2.8315 & \textbf{0.0018} & 0.0043 & 5.1773 & \textbf{0.0006} \\
& & 5  & 0.1179 & 0.2089 & \textbf{0.0373} & 0.0487 & 0.3862 & \textbf{0.0018} & 0.0042 & 0.0751 & \textbf{0.0006} \\
& & 15 & 0.1169 & 0.1301 & \textbf{0.0310} & 0.0421 & 0.0826 & \textbf{0.0014} & 0.0039 & 0.0135 & \textbf{0.0006} \\
& & 25 & 0.1140 & 0.1615 & \textbf{0.0288} & 0.0367 & 0.1069 & \textbf{0.0013} & 0.0036 & 0.0179 & \textbf{0.0006} \\
\cmidrule(lr){2-12}
& \multirow{4}{*}{25\%}
& 0  & 0.0821 & 0.2347 & \textbf{0.0315} & 0.0416 & 0.2279 & \textbf{0.0017} & 0.0032 & 0.2615 & \textbf{0.0006} \\
& & 5  & 0.0848 & 0.1377 & \textbf{0.0301} & 0.0387 & 0.1153 & \textbf{0.0016} & 0.0031 & 0.0202 & \textbf{0.0006} \\
& & 15 & 0.0843 & 0.1583 & \textbf{0.0265} & 0.0336 & 0.1214 & \textbf{0.0014} & 0.0029 & 0.0208 & \textbf{0.0006} \\
& & 25 & 0.0836 & 0.1691 & \textbf{0.0246} & 0.0294 & 0.1170 & \textbf{0.0012} & 0.0027 & 0.0210 & \textbf{0.0005} \\
\midrule
\multirow{12}{*}{M3}
& \multirow{4}{*}{5\%}
& 0  & 0.2662 & 0.4920 & \textbf{0.1833} & 0.1463 & 0.2962 & \textbf{0.0289} & 0.0028 & 0.2066 & \textbf{0.0008} \\
& & 5  & 0.2466 & 0.1579 & \textbf{0.1352} & 0.1356 & 0.0697 & \textbf{0.0124} & 0.0030 & 0.0035 & \textbf{0.0008} \\
& & 15 & 0.2275 & 0.1171 & \textbf{0.0980} & 0.1167 & 0.0420 & \textbf{0.0071} & 0.0033 & 0.0028 & \textbf{0.0008} \\
& & 25 & 0.2117 & 0.1049 & \textbf{0.0789} & 0.1009 & 0.0487 & \textbf{0.0039} & 0.0034 & 0.0030 & \textbf{0.0008} \\
\cmidrule(lr){2-12}
& \multirow{4}{*}{15\%}
& 0  & 0.1187 & 0.2082 & \textbf{0.0372} & 0.0523 & 0.2389 & \textbf{0.0017} & 0.0043 & 0.3609 & \textbf{0.0006} \\
& & 5  & 0.1179 & 0.1511 & \textbf{0.0347} & 0.0487 & 0.1607 & \textbf{0.0016} & 0.0042 & 0.0242 & \textbf{0.0006} \\
& & 15 & 0.1169 & 0.1404 & \textbf{0.0305} & 0.0421 & 0.1103 & \textbf{0.0014} & 0.0039 & 0.0220 & \textbf{0.0006} \\
& & 25 & 0.1140 & 0.1466 & \textbf{0.0289} & 0.0367 & 0.0948 & \textbf{0.0013} & 0.0036 & 0.0173 & \textbf{0.0006} \\
\cmidrule(lr){2-12}
& \multirow{4}{*}{25\%}
& 0  & 0.0821 & 0.2008 & \textbf{0.0316} & 0.0416 & 0.1948 & \textbf{0.0017} & 0.0032 & 0.3363 & \textbf{0.0006} \\
& & 5  & 0.0848 & 0.1888 & \textbf{0.0296} & 0.0387 & 0.1965 & \textbf{0.0016} & 0.0031 & 0.0331 & \textbf{0.0006} \\
& & 15 & 0.0843 & 0.1616 & \textbf{0.0264} & 0.0336 & 0.1274 & \textbf{0.0014} & 0.0029 & 0.0254 & \textbf{0.0006} \\
& & 25 & 0.0836 & 0.1861 & \textbf{0.0245} & 0.0294 & 0.1309 & \textbf{0.0013} & 0.0027 & 0.0245 & \textbf{0.0006} \\
\bottomrule
\end{tabular}
\end{table}

Tables \ref{tab:full_result} report LPIPS, Masked-MSE, and Sobel-MSE across all combinations of training mode, observation density (row-mask: $5 \% \rightarrow 25 \%$), probe intensity ($\lambda: 0 \rightarrow 25$), and sampler (AAS-only, RePaint, PMA-Diffusion). To interpret these results, it is useful to recall how visibility is defined across different models. In the \textbf{Full-obs model} (M1), the network is trained on fully observed (ground-truth) speed fields, and the row visibility and $\lambda$ values apply only at sampling stage. In contrast, the \textbf{Single-mask} (M2) and \textbf{Double-mask} (M3) models are both trained and evaluated under the same partial-observation condition. Thus, for M2 and M3, the specified row and $\lambda$ values reflect the entire modeling pipeline, not only the inference stage.

We organize the discussion in three steps. First, we fix the training strategy to the Full-obs model and compare the three sampling strategies to assess the value of the proposed PMA-Diffusion sampler. Second, we fix the sampler to PMA-Diffusion and compare the three training strategies to evaluate how close mask-aware training can get to the Full-obs upper bound. Finally, we summarize the overall trends and additional analyses.

\begin{figure}[width=.99\linewidth,cols=4,pos=t]
    \centering
    \includegraphics[width=0.99\linewidth]{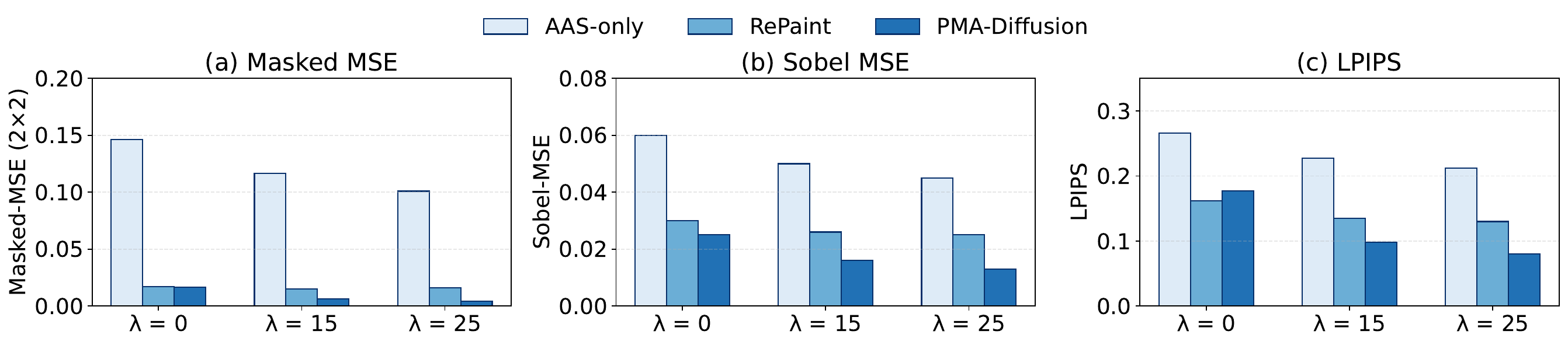}
    \caption{Performance of different samplers based on Full-obs model ($row = 5\%$).}
     \label{fig:full_obs_masked_mse}
\end{figure}

\subsubsection{Effect of sampling strategies under a fully observed prior}
\label{sec:exp_sampling_fullobs}
We first consider the Full-obs prior (M1), which is trained on fully observed speed fields and therefore represents the best-case scenario for the diffusion architecture itself. This setting isolates the effect of the sampling strategy, since differences cannot be attributed to missing data during training.

Figure~\ref{fig:full_obs_masked_mse} shows Masked-MSE, Sobel MSE, and LPIPS for the three samplers under the sparse setting with $5\%$ fixed rows and no probe vehicles. In this case, the AAS already achieves a Masked-MSE of $0.1463$ on the unobserved pixels. RePaint further reduces reduces the error to $0.0172$ by iteratively mixing forward and reverse steps. Adding the physics projector (PMA-Diffusion) yields a slight additional reduction in Masked-MSE and improves Sobel-MSE, indicating better alignment of wave fronts.

Across the other row and $\lambda$ combinations for M1 in Table~\ref{tab:full_result}, the same pattern holds. AAS sampling provides a reasonable baseline. RePaint consistently improves both Masked-MSE and LPIPS by exploiting iterative resampling. The proposed PMA-Diffusion sampler, which augments RePaint with the AAS projector, either matches or improves on RePaint in almost all settings and is particularly effective at reducing Sobel-MSE in sparse configurations. This confirms that, when the prior is well-trained, the PMA-Diffusion sampler gives the best overall reconstruction quality among the three options.

\subsubsection{Effect of training strategies under the PMA-Diffusion sampler}
\label{sec:exp_training_pma}
Based on the comparison above, we fix the sampler to PMA-Diffusion and compare the three training strategies: Full-obs (M1), Single-mask (M2), and Double-mask (M3). This addresses RQ1 by asking how much accuracy is lost when the prior is trained only on incomplete fields, and to what extent the proposed double-mask strategy can close the gap to the Full-obs upper bound.

\begin{figure}[width=.99\linewidth,cols=4,pos=t]
    \centering
    \includegraphics[width=0.9\linewidth]{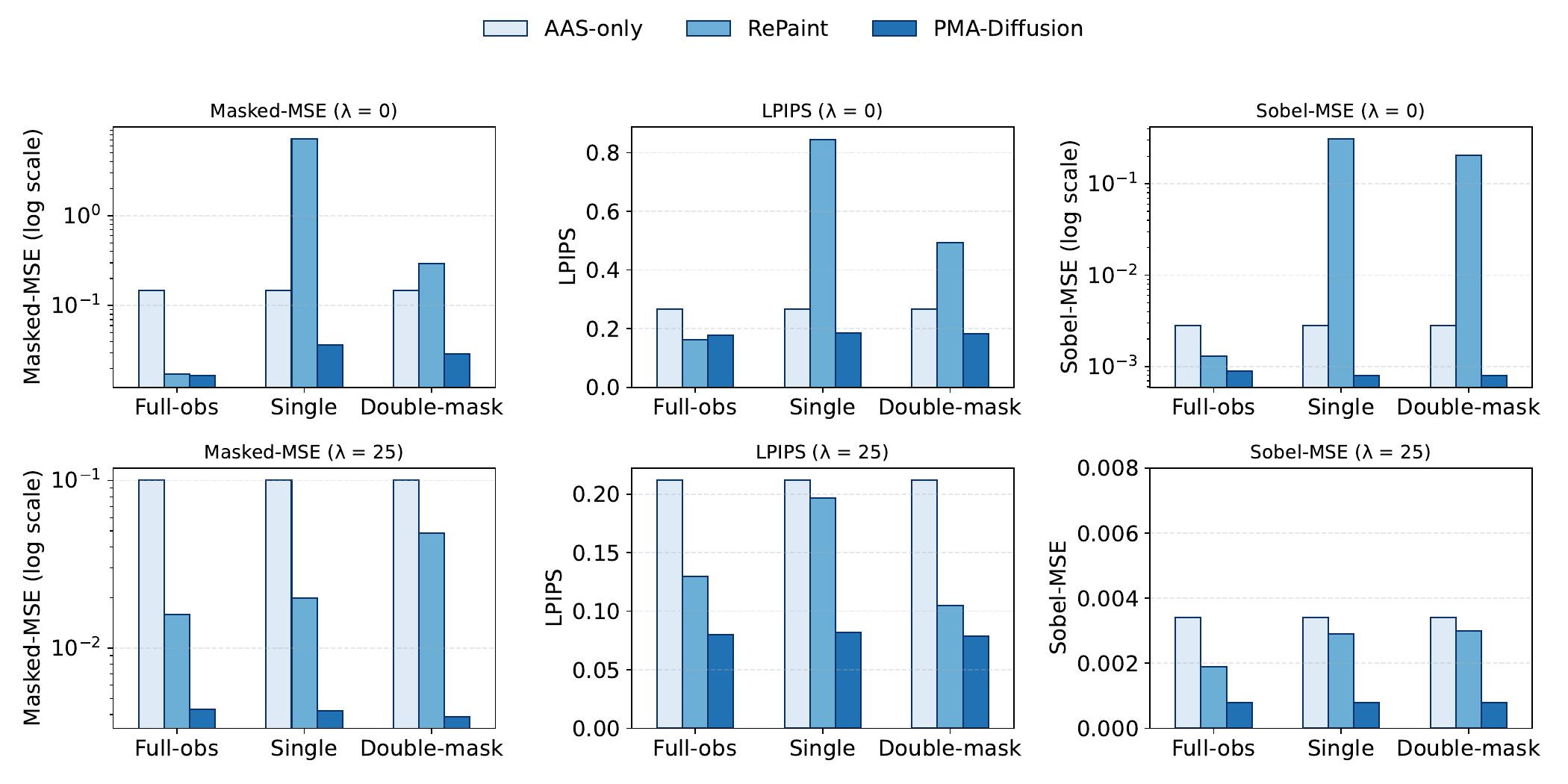}
    \caption{Quantitative comparison of Masked-MSE, LPIPS, and Sobel-MSE across three training modes (Full-obs, Single-mask, Double-mask) under Row = 5\% for two probe intensities, $\lambda = 0$ and $\lambda = 25$.}
    \label{fig:row5_two_panel}
\end{figure}

Under the most challenging setting with $5\%$ rows and $\lambda=0$, the differences between the three training strategies are clear. With PMA-Diffusion sampling, the Full-obs model achieves a Masked-MSE of $0.0167$, while the Single-mask model reaches $0.0364$ and the Double-mask model reaches $0.0289$ (Table~\ref{tab:full_result}). In the same setting, LPIPS follows the same ordering, and Sobel-MSE remains at the $10^{-3}$ level for all three, showing that all models can reproduce wave structures but differ in low-frequency accuracy. Figure~\ref{fig:row5_two_panel} summarizes Masked-MSE trends across $\lambda$ for the most sparse row setting.

The Single-mask prior suffers whenever parts of the grid are rarely or never visible during training. In this regime, the RePaint samplers can be unstable: for example, with $5\%$ rows and $\lambda=0$, the RePaint sampler yields a Masked-MSE above $7$ and a LPIPS of $0.8455$ for M2, reflecting severe over-extrapolation in unobserved regions. Once PMA-Diffusion is used, these pathologies are largely corrected and the error drops by several orders of magnitude, but the gap to the Full-obs prior remains noticeable.

The Double-mask prior (M3) substantially improves robustness while using the same underlying dataset as M2. By randomly hiding additional pixels during training, it forces the model to learn to interpolate across persistently unseen regions. As a result, under $5\%$ rows and $\lambda=0$, the Double-mask model with PMA-Diffusion reduces Masked-MSE from $0.0364$ (M2) to $0.0289$ (M3), and in many denser settings the gap to the Full-obs prior becomes small. For example, for $15\%$ rows and $\lambda=15$, the Full-obs model with PMA-Diffusion has Masked-MSE $0.0016$, while the Double-mask counterpart achieves $0.0014$. In some dense states (e.g., $25\%$ rows and $\lambda=25$), the Single-mask and Double-mask models with PMA-Diffusion both approach the Full-obs baseline and can slightly outperform each other depending on the metric, suggesting that when every pixel is frequently observed in historical data, Single-mask training is sufficient.

Under the PMA-Diffusion sampler, the proposed Double-mask training strategy close most of the gap to the Full-obs upper bound, especially in sparse states. Single-mask training strategy can approached the Full-obs performance only when each location is observed with sufficient frequency in the training data. 
Also, physics guidance through the AAS projector maintains performance across all configurations in Table~\ref{tab:full_result}. It either leaves the metrics unchanged or improves them. The gain is most significant under sparse observations (e.g., $5\%$ rows), where the projector stabilizes the unobserved regions.

\subsubsection{Qualitative illustration}
\begin{figure}[width=.99\linewidth,cols=4,pos=t]
    \centering
    \begin{subfigure}[t]{0.3\textwidth}
        \centering
        \includegraphics[width=\textwidth]{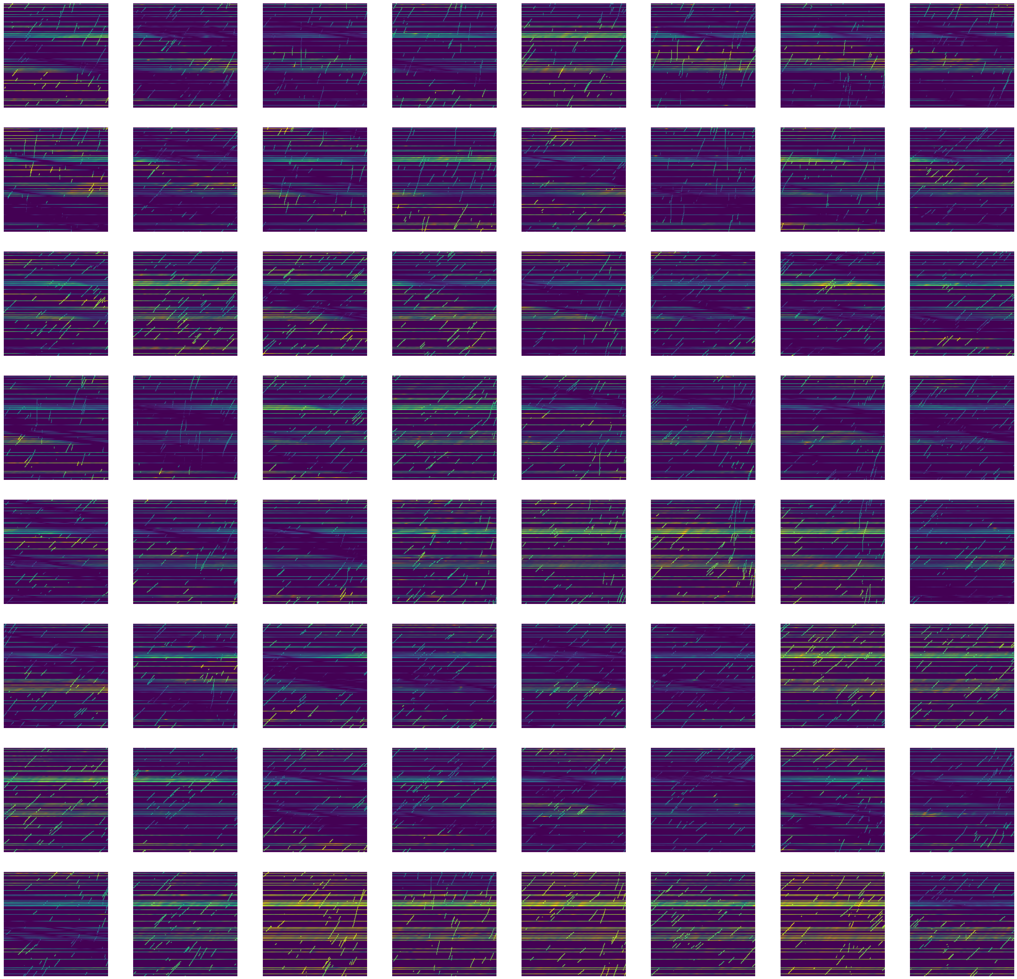}
        \centerline{(a) Observed regions (low-obs)}
    \end{subfigure}
    \hspace{0.02\textwidth}
    \begin{subfigure}[t]{0.3\textwidth}
        \centering
        \includegraphics[width=\textwidth]{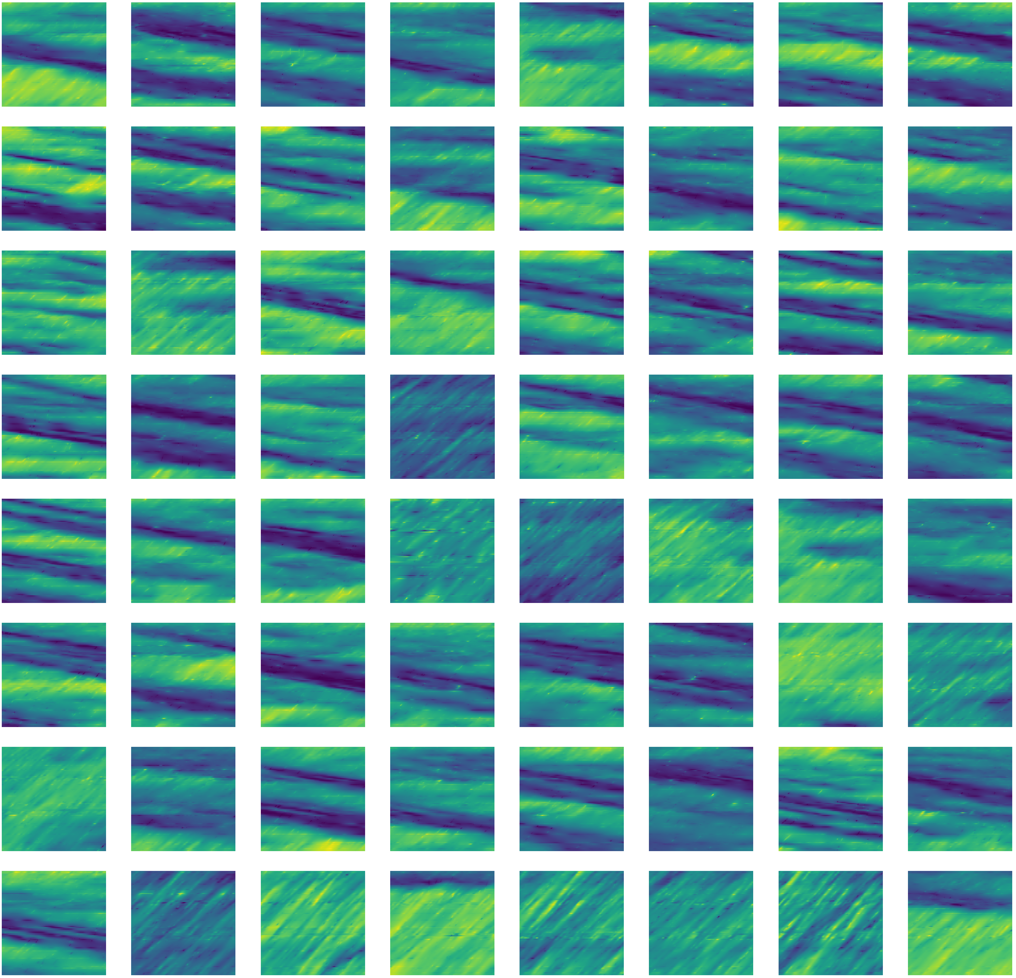}
        \centerline{(b) PMA-Diffusion (low-obs)}
    \end{subfigure}
    \hspace{0.02\textwidth}
    \begin{subfigure}[t]{0.3\textwidth}
        \centering
        \includegraphics[width=\textwidth]{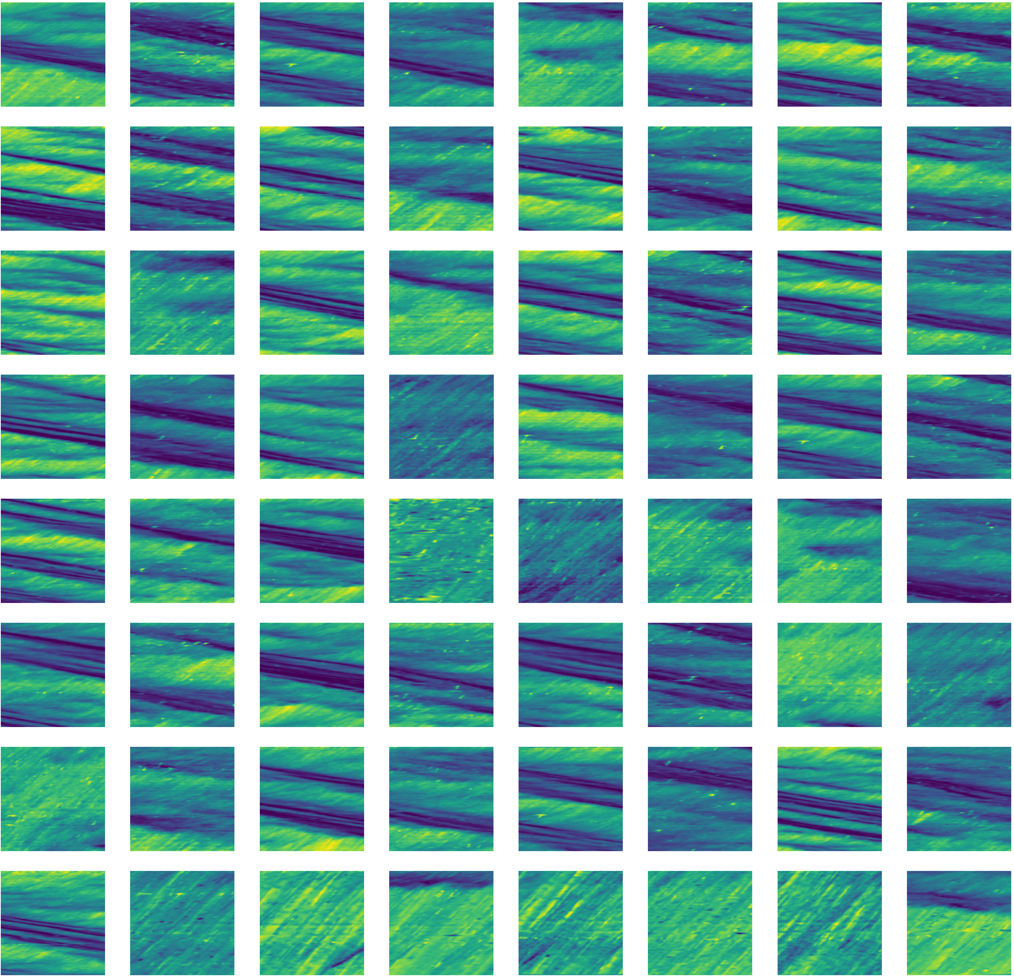}
        \centerline{(c) Ground truth (low-obs)}
    \end{subfigure}

    \vspace{0.02\textwidth}

    \begin{subfigure}[t]{0.3\textwidth}
        \centering
        \includegraphics[width=\textwidth]{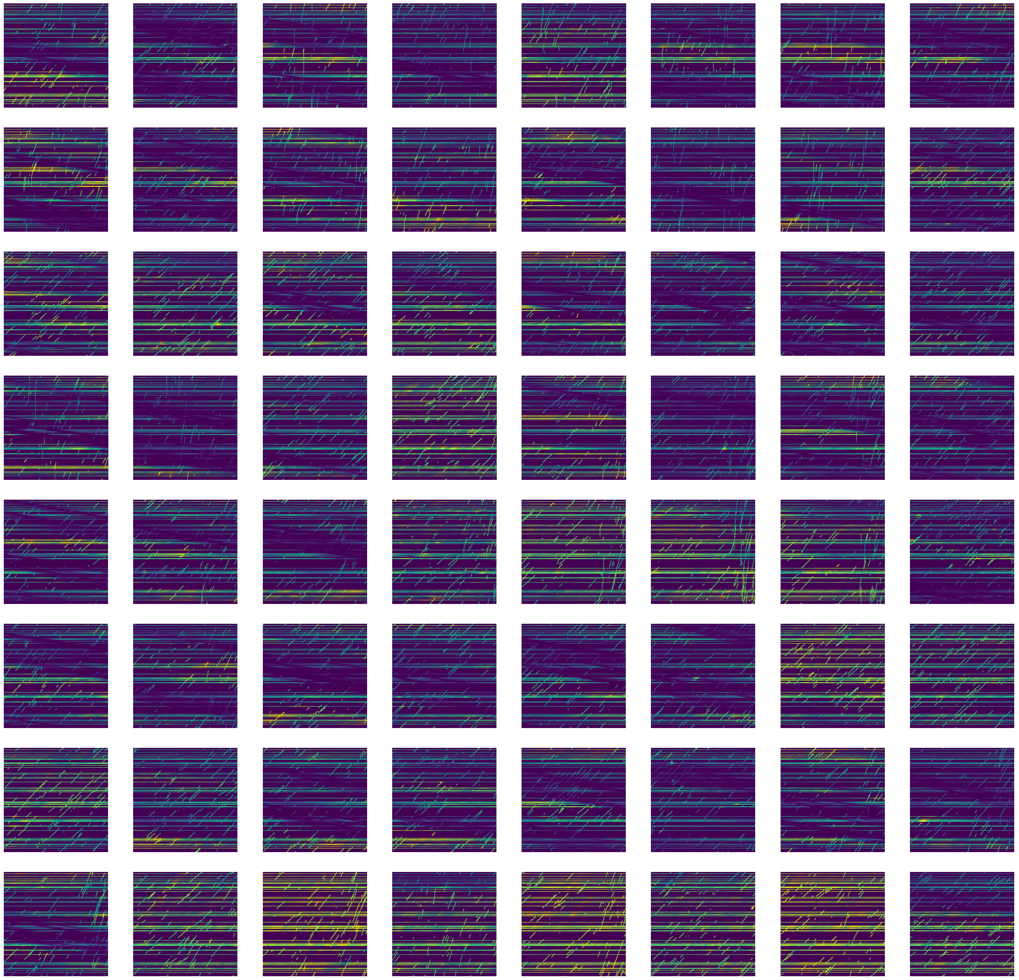}
        \centerline{(d) Observed regions (high-obs)}
    \end{subfigure}
    \hspace{0.02\textwidth}
    \begin{subfigure}[t]{0.3\textwidth}
        \centering
        \includegraphics[width=\textwidth]{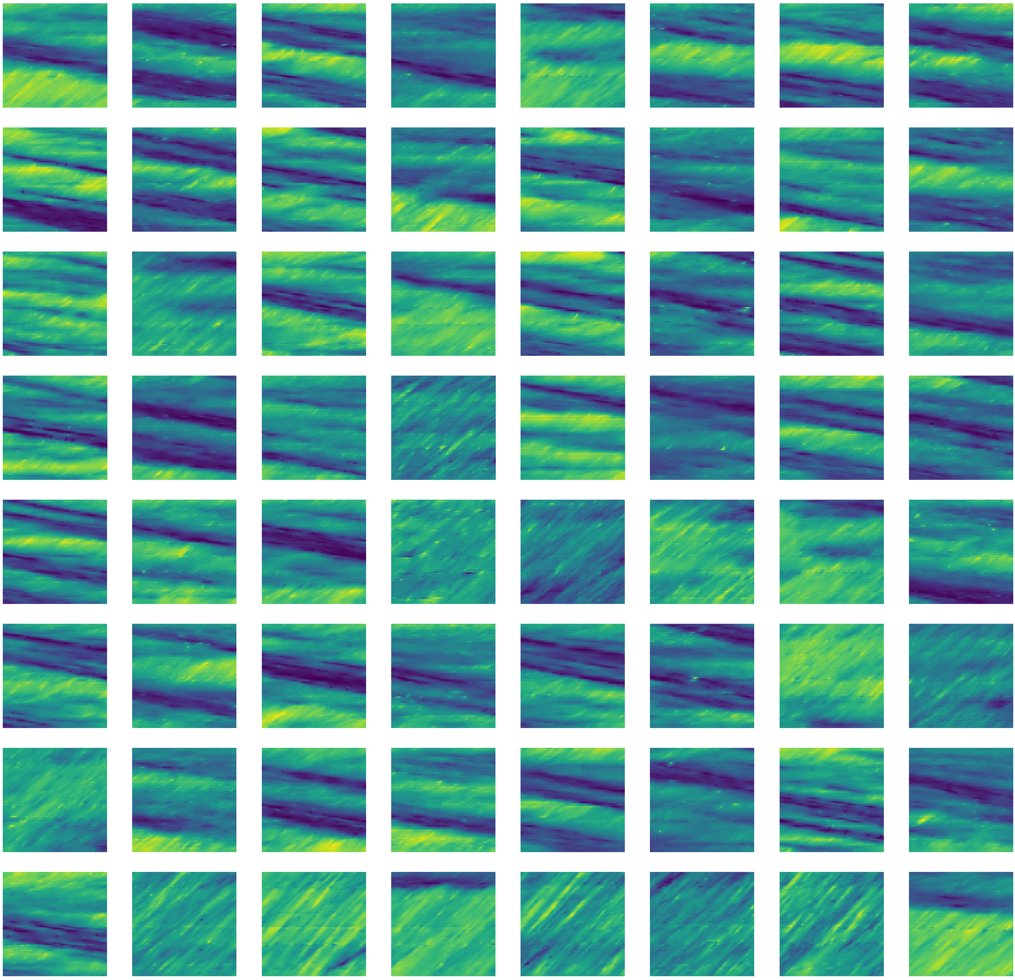}
        \centerline{(e) PMA-Diffusion (high-obs)}
    \end{subfigure}
    \hspace{0.02\textwidth}
    \begin{subfigure}[t]{0.3\textwidth}
        \centering
        \includegraphics[width=\textwidth]{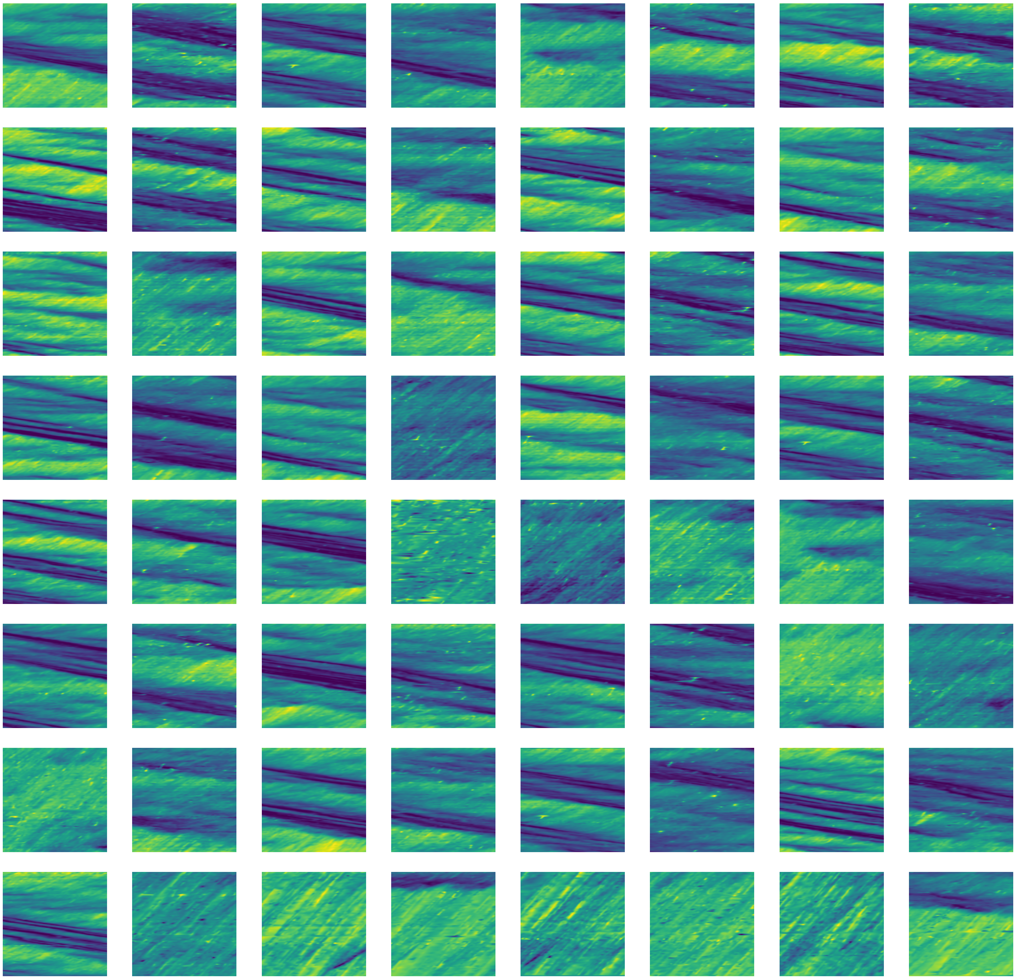}
        \centerline{(f) Ground truth (high-obs)}
    \end{subfigure}

    \caption{Visual comparison between Observed regions, PMA-Diffusion, and Ground truth under the Double-mask model for low-observation ($\text{row}=15\%$, $\lambda=15$) and high-observation ($\text{row}=25\%$, $\lambda=25$) scenarios.}
    \label{fig:qualitative}
\end{figure}

Figure~\ref{fig:qualitative} compares the PMA-Diffusion reconstruction against the ground truth using the Double-mask model under both low-observation ($\text{row}=15\%$, $\lambda=15$) and high-observation ($\text{row}=25\%$, $\lambda=25$) settings.
While the generated speed field appears smoother and exhibits slight discontinuities between generated and observed regions, its overall spatiotemporal structure remains consistent and physically coherent. This suggests that the model effectively captures macroscopic traffic patterns, even if some high-frequency details are attenuated during sampling. 

\begin{table}[width=.99\linewidth,cols=4,pos=t]
\centering
\scriptsize
\setlength{\tabcolsep}{4pt}
\caption{Quantitative results for the double-mask model (M3) on $64{\times}64$ grids under different row coverage levels (Row) and probe intensities ($\lambda$). For each setting, we report LPIPS, Masked-MSE (2$\times$2), and Sobel-MSE for three reconstruction methods (AAS-only, RePaint, and PMA-Diffusion). All metrics are computed on unobserved pixels, and lower values indicate better reconstruction quality.}
\label{tab:64_summary}
\begin{tabular}{llc rrr rrr rrr}
\toprule
\multirow{2}{*}{Mode} & \multirow{2}{*}{Row} & \multirow{2}{*}{$\lambda$} &
\multicolumn{3}{c}{LPIPS} &
\multicolumn{3}{c}{Masked-MSE (2$\times$2)} &
\multicolumn{3}{c}{Sobel-MSE} \\
\cmidrule(lr){4-6}\cmidrule(lr){7-9}\cmidrule(l){10-12}
& & & AAS-only & RePaint & PMA & AAS-only & RePaint & PMA & AAS-only & RePaint & PMA \\
\midrule
\multirow{11}{*}{Double-mask model (M3)}
& \multirow{4}{*}{5\%}
& 0  & 0.2333 & 0.9164 & 0.0719 & 0.4122 & 0.2194 & 0.0120 & 0.0013 & 0.5189 & 0.0009 \\
& & 5  & 0.1776 & 0.0479 & 0.0124 & 0.2706 & 0.1184 & 0.0032 & 0.0012 & 0.0017 & 0.0009 \\
& & 15 & 0.0863 & 0.0388 & 0.0050 & 0.1306 & 0.0541 & 0.0020 & 0.0010 & 0.0019 & 0.0008 \\
& & 25 & 0.0536 & 0.0676 & 0.0035 & 0.0746 & 0.0957 & 0.0018 & 0.0009 & 0.0039 & 0.0007 \\
\cmidrule(lr){2-12}
& \multirow{4}{*}{15\%}
& 0  & 0.0388 & 0.3976 & 0.0032 & 0.0808 & 1.4737 & 0.0018 & 0.0014 & 1.6367 & 0.0008 \\
& & 5  & 0.0340 & 0.0323 & 0.0030 & 0.0548 & 0.0609 & 0.0017 & 0.0011 & 0.0053 & 0.0007 \\
& & 15 & 0.0232 & 0.0278 & 0.0026 & 0.0283 & 0.0238 & 0.0016 & 0.0008 & 0.0013 & 0.0007 \\
& & 25 & 0.0147 & 0.0219 & 0.0024 & 0.0171 & 0.0158 & 0.0015 & 0.0006 & 0.0010 & 0.0005 \\
\cmidrule(lr){2-12}
& \multirow{3}{*}{25\%}
& 0  & 0.0399 & 0.2450 & 0.0032 & 0.0745 & 1.6512 & 0.0017 & 0.0007 & 0.1234 & 0.0007 \\
& & 5  & 0.0325 & 0.0606 & 0.0028 & 0.0496 & 0.2309 & 0.0016 & 0.0007 & 0.0086 & 0.0006 \\
& & 15  & 0.0226 & 0.0408 & 0.0026 & 0.0262 & 0.0598 & 0.0014 & 0.0006 & 0.0021 & 0.0006 \\
\bottomrule
\end{tabular}
\end{table}

\subsubsection{Performance across different Spatial-Temporal Resolutions}
To examine scalability and resolution sensitivity, we re-discretize the same I-24 dataset from $200{\times}200$ grids ($\Delta t{=}1$\,s, $\Delta x{=}100$\,ft) to $64{\times}64$ grids ($\Delta t{=}5$\,s, $\Delta x{=}200$\,ft), while strictly preserving the spatiotemporal domain and visibility patterns. 
Under the $64{\times}64$ setting, we use the same model family and training protocol: a UNet with one input channel and 64 base feature channels, three resolution levels with channel multipliers $(1,2,4)$, and 4-head linear attention in the bottleneck; a linear $\beta$-schedule with $T{=}500$ steps ($\beta_0{=}10^{-4}$, $\beta_T{=}2{\times}10^{-2}$); Huber loss ($\delta{=}1$); and Adam (learning rate $5{\times}10^{-4}$, batch size $=250$). Training runs for 10 epochs (approximately $1.8\times10^{3}$ iterations), requiring about 15 minutes of GPU time. During inference, we apply the same four sampling methods as in the high-resolution setting.

The results show that the trends observed at $64{\times}64$ are consistent with those at $200{\times}200$, and are numerically even more stable. PMA-Diffusion achieves the best performance across the majority of configurations. For example, under the Double-mask (M3) setting with Row = $15\%$ and $\lambda = 15$, the Masked-MSE decreases from $0.0238$ (RePaint) to $0.0016$ after applying AAS. Correspondingly, LPIPS improves from $0.0278$ to $0.0026$, and Sobel-MSE improves from $0.0013$ to $0.0007$. 

In the most sparse setting (Row = $5\%$, $\lambda = 0$), both the Single-mask and Double-mask priors can exhibit instability or over-extrapolation when using RePaint, whereas PMA-Diffusion reduces Masked-MSE to $0.0120$ (M3), representing an order-of-magnitude improvement over RePaint ($0.2194$). Even in the Full-obs model (M1), PMA-Diffusion achieves a Masked-MSE of $0.0013$ at Row = $25\%$, $\lambda = 25$, indicating that posterior refinement guided by physics remains beneficial even when the prior is already strong. For brevity, Table~\ref{tab:64_summary} reports representative configurations on $64{\times}64$ grids

\section{Conclusion}
\label{sec:conclu}
This paper presents \textbf{PMA-Diffusion}, a physics-guided, mask-aware diffusion framework for highway traffic state estimation from sparse loop detector and probe vehicle data. The framework reconstructs unobserved speed fields from incomplete observations by combining a diffusion prior with a physics-guided posterior sampler. The prior is trained directly on incompletely observed speed fields using two masking strategies: a single-mask strategy that follows the actual sensor coverage, and a double-mask strategy that adds an auxiliary mask to temporarily hide always-visible pixels. At inference, the sampler alternates reverse-diffusion updates, an observation projection that enforces data fidelity on the visible entries, and a physics-guided projector based on adaptive anisotropic smoothing that enforces the missing region toward kinematic-wave structure.

The study addresses two main questions: How well a diffusion prior can be learned on incomplete speed fields, and given a fixed prior, how much the sampling stage can recover under varying sensing sparsity. The experiments on the I-24 MOTION data show that a mask-aware diffusion prior trained with the double-mask strategy closes most of the gap to a fully supervised model, especially when some pixels are rarely or never observed. Even under severe sparsity (e.g., $5\%$ visibility), the combination of a mask-aware prior and the physics-guided sampler substantially improves reconstruction quality compared with both baselines, and achieves near-full-observation accuracy in several denser settings. These results indicate that PMA-Diffusion provides a practical way to learn reusable high-resolution traffic speed priors from incomplete data and to adapt them to different sensing layouts through our physics-guided posterior sampler.

Several assumptions made in this study limit the current model. Probe vehicle and loop detector data are treated as time-aligned (Assumption~\ref{asm:latency}), and low-frequency sensor bias is assumed to be small and smooth enough to be absorbed into Gaussian noise (Assumption~\ref{asm:drift}), leading to the simplified observation model in Eq.~\eqref{eq:obs-simple}. In addition, the current physics projector acts only on speeds and does not explicitly incorporate other traffic-flow constraints.

Future work can relax these assumptions and better account for heterogeneous data sources. One direction is to develop diffusion training and sampling schemes that model modality-specific noise and bias for loop detectors and probe vehicles, rather than absorbing all differences into a single Gaussian term. This includes handling non-negligible latency, sensor drift and non-Gaussian noise, and possible miscalibration, and examining how these factors affect the diffusion prior and posterior sampling. Another direction is to extend the physics module beyond adaptive anisotropic smoothing. The same projection interface can integrate additional traffic-flow constraints, such as conservation laws, fundamental-diagram consistency, or network-level macroscopic structure. 


\section*{Declaration of AI and AI-assisted technologies in the writing process}
During the preparation of this work, the authors used AI-assisted language polishing tools for language assistance. After utilizing these tools, the authors thoroughly reviewed and revised the content as necessary and take full responsibility for the final version of the published article.

\section*{Acknowledgement}
We acknowledge the CSE Data Science Initiative at the University of Minnesota College of Science and Engineering for funding through the ADC Graduate Assistantship.

\printcredits

\bibliographystyle{cas-model2-names}
\bibliography{cas-sc-template}

\end{document}